\documentclass{article}

\PassOptionsToPackage{numbers}{natbib}
\usepackage[preprint]{neurips_2026}
\newcommand{\rev}[1]{#1}
\def\bibpath{references}
\def\bibsize{}

\usepackage[utf8]{inputenc}
\usepackage[T1]{fontenc}
\usepackage{hyperref}
\usepackage{xurl}
\usepackage{booktabs}
\usepackage{amsfonts}
\usepackage{amsmath}
\usepackage{nicefrac}
\usepackage{microtype}
\usepackage{xcolor}
\usepackage{graphicx}
\usepackage{enumitem}
\usepackage{etoolbox}

\title{Towards a Deterministic Math Solver for Clinical Language Models}

\author{
  Felipe Ocampo Osorio \\ MIT Critical Data \\ \texttt{felipeos@mit.edu}
  \And
  Sebastián Andrés Cajas Ordóñez \\MIT Critical Data \\ \texttt{sebasmos@mit.edu}
  \And
  Maximin Lange \\ MIT Critical Data \\ \texttt{mlange2@mit.edu}
  \And
  Rafi Al Attrach \\ MIT Critical Data \\ \texttt{rafiaa@mit.edu}
  \And
  Sahil Kapadia \\ UNC Chapel-Hil \\ \texttt{sahilk@unc.edu}
  \And
  Zakaria Laouabdia Sellami \\ University of Pavia \\ \footnotesize\texttt{zakaria.laouabdiasella01@universitadipavia.it}
  \And
  Angelo Antonio Talio \\ Humanitas University \\ \texttt{angelo.talio@humanitas.it}
  \And
  Leo Anthony Celi \\ MIT Critical Data \\ \texttt{lceli@mit.edu}
}

\begin{document}

\maketitle

\begin{abstract}
Large language models are unreliable at arithmetic, which is a problem for clinical calculators where a single numerical error changes the recommendation. The standard response is to hardcode each calculator as a validated function, one at a time. We test an alternative: the model does not calculate. Instead, it writes case-specific Python that a restricted local executor runs as a deterministic solver, and the model's task reduces to deciding how to use it. We evaluate this Program-Solve interface on MedCalc-Bench Verified (1,100 cases, 55 calculators) against direct model arithmetic and a hand-written 22-calculator library, using Qwen2.5-7B and Qwen2.5-32B-AWQ, after auditing the benchmark's formulas against current clinical guidelines and flagging 16 of 55 with version, use or coefficient concerns. With formulas and gold variables supplied and both routes reading the whole note, handing off to the solver is not a reliable advantage at 7B (75.31\% against 72.02\%, a paired $+3.29$ points with a 95\% calculator-cluster interval of $[-3.49, 10.38]$) but is one at 32B (90.53\% against 83.47\%, $+7.05$ $[0.47, 14.60]$, clear of zero). The hand-written library is exact on its 440 supported cases but abstains elsewhere (40.0\% overall). Adding an executor thus helps some open-weight models more than others even under matched formula, variable and note access, and is not a substitute for verified formulas or reliable variable extraction either way.
Code and data: \url{https://github.com/felipeocampoos/Towards-a-Deterministic-Math-Solver-for-Clinical-Language-Models}.
\end{abstract}

\section{Introduction}
\label{sec:intro}

Automating clinical calculators such as APACHE II \cite{knaus1985apache} with a language model
requires selecting the formula, extracting variables from a free-text note and executing the
arithmetic, on which MedCalc-Bench \cite{khandekar2024medcalc} shows models losing accuracy: Goodell
and colleagues report incorrect answers in about one third of unaided ChatGPT trials across 48
calculation tasks \cite{goodell2025large}. The standard remedy is a hand-written function per calculator, each written, validated and
maintained, with every case outside the set unanswerable: the evaluated 22-calculator library
implements 440 of 1{,}100 cases, 40.0\% full-set accuracy under abstention. We evaluate whether calculator-specific execution can be replaced by a general interface: the model
is given the case and writes a short Python program, and a restricted executor runs it and returns a
number, a date or a gestational-age tuple. The executor contains no calculator-specific functions or
constants; the model translates the supplied clinical formula into code. Clinically, execution
fidelity is necessary but not sufficient: the formulas a calculator encodes are themselves
versioned, and several in routine use (race-free eGFR, MELD 3.0, PREVENT, the Sampson LDL equation)
have replaced predecessors that benchmarks may still reward. Local serving avoids
external API calls and data transmission; its unmeasured costs are in Section~\ref{sec:limitations}.

\paragraph{Related work and contribution.} Program-aided reasoning has the model emit code for an
interpreter \cite{gao2023pal, chen2023program}; chain-of-thought prompting \cite{wei2022chain} is
the in-context alternative. Executing that code carries a security exposure distinct from whether
it is correct \cite{wang2024codeact}. MedCalc-Bench formalised calculator invocation
\cite{khandekar2024medcalc}; MedRaC pairs retrieval with Python execution and
scores formula selection, extraction and arithmetic separately \cite{wang2025scores}; RiskAgent
selects among validated tools \cite{liu2025riskagent}; a clinical-calculator chatbot routes to
verifiable calculators \cite{kumar2024chatbot}; MeNTi bridges calculators and agents through nested
tool calling \cite{zhu2024menti}; verifiable-reward training raises
the aggregate \cite{wangmedcalc}; decomposition adds failure points when extraction is incomplete
\cite{shah2024accuracy}; most calculator-selection errors are comprehension errors, not arithmetic
ones \cite{wan2025humans}. A code-interpreter arm compared against task-specific calculator tools
found the tools more accurate \cite{goodell2025large}. AgentMD automates the tool curation we
describe as a maintenance burden \cite{jin2025agentmd}, and the coverage-accuracy trade-off a
partial library exhibits is the abstention problem \cite{xin2021art, wen2025know}. Our
contribution is a controlled comparison of
case-specific program generation against direct arithmetic and the hand-written alternative, under
matched formula, variable and note access. \rev{It establishes what execution does and does not fix on
one benchmark; generalization to unseen formulas, languages or settings is outside its scope.}

\begin{figure}[!t]
% Built by figures/graphical_abstract/build_graphical_abstract.py in the code repo from the committed
% per-case CSVs (five seeds, 1,100 cases). Rerun that script and copy the PDF here; nothing is typed in.
% 2026-09-04: reduced-text build (labels only, larger fonts, 660x312 canvas); the caption carries the explanation.
\centering
\IfFileExists{graphical_abstract.pdf}{%
\includegraphics[width=0.90\linewidth]{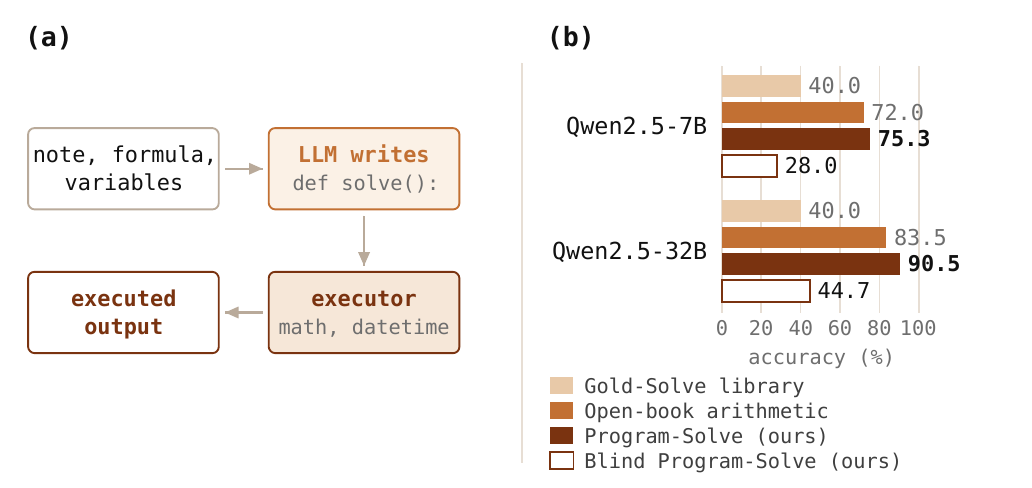}%
}{%
\fbox{\parbox[c][0.40\linewidth][c]{0.85\linewidth}{\centering
Graphical abstract unavailable.}}%
}
\caption{\textbf{(a)} The model writes a program from note, formula and variables and an executor
without calculator code runs it: all 55 calculators attempted, the library implements 22.
\textbf{(b)} Accuracy on 1{,}100 cases; open bar: Blind Program-Solve. Program-Solve minus Open-book
arithmetic: calculator-cluster interval (Table~\ref{tab:gaps}) crosses zero at 7B and clears it at
32B.}
\label{fig:abstract}
\end{figure}

\section{Method}
\label{sec:method}

\paragraph{Data and models.} MedCalc-Bench Verified, 1{,}100 test cases across 55 calculators,
scored under benchmark-defined tolerance. Two open-weight models, Qwen2.5-7B-Instruct (bf16) and
Qwen2.5-32B-Instruct-AWQ (4-bit), served through vLLM on cloud H100 GPUs (tensor-parallel 1, one
H100 each; other checkpoints likewise one H100 or, for Mistral, two under tensor parallelism), at
five seeds (42 to
46) over all 1{,}100 cases. The
worked one-shot example comes from the benchmark's separate one-shot split; no test case's gold
answer or explanation enters any prompt. MedCalc-Bench Verified is CC-BY-SA~4.0, both models
Apache~2.0.

\paragraph{Executor.} A fresh subprocess per program: a builtins allow-list with no file, eval or
exec primitives; imports limited to the standard math, date, time and calendar modules; static
rejection of async and generator constructs; 256\,MB and CPU limits, 5\,s wall clock, an
executed-line cap. \rev{The executor is restricted but is no sandbox: there is no container or syscall filter.}

\paragraph{Arms.} Table~\ref{tab:ladder} states each arm's formula access, variable access and
execution method. \textbf{Note only} gets the note. \textbf{Page without formula} adds the calculator's page with the formula suppressed and
\textbf{Open-book arithmetic} adds the formula, both with variables given and the model doing the
arithmetic. \textbf{Extract-Solve library} extracts variables and calls one of 22 hand-written
Python calculators; \textbf{Gold-Solve library} gives those same calculators the gold variables.
\rev{The 22 were not chosen by a stated criterion: all are laboratory, physical or date calculators and
none is a point-based score, an opportunistic set rather than a principled one. Because the
benchmark is balanced at 20 cases per calculator, a library's coverage is a fixed function of how
many calculators it implements, so its full-set accuracy under abstention is bounded by that count
(Figure~\ref{fig:coverage}).}
\textbf{Program-Solve} is ours: the model writes a program that the executor runs,
given the formula text and gold variables, the inputs Open-book arithmetic receives;
\textbf{Blind Program-Solve} gets neither and extracts its own variables. Calculator support,
attempted-answer rate, valid-program rate and correct-answer rate are distinct measures; Table~\ref{tab:ladder}
reports correct-answer rates only.

\paragraph{Comparison design.} Comparisons are distinguished by formula access, variable access and
note length. Note only (120 words) and Page without formula (250 words) cap the note; Open-book
arithmetic, Program-Solve, Blind Program-Solve and the Extract-Solve library all read the
whole note, so Program-Solve against Open-book arithmetic is matched on formula, variables and note
length. A 250-word one-shot arm without gold variables
(Table~\ref{tab:s16-entitynone}) separates budget from access: with variable access fixed, the
larger budget alone adds 4 to 12 points in every family. Program-Solve against the Extract-Solve library is not input-matched; the matched pairs are
Program-Solve/Gold-Solve library and Blind Program-Solve/Extract-Solve library. Decoding, token,
note-budget and executor settings are in Table~\ref{tab:s14-settings}; every arm's protocol is in
Appendix~\ref{sec:protocols}.

\paragraph{Statistics.} Cases nest in 55 calculators and
recur across five seeds, so the primary uncertainty is a cluster bootstrap over calculators
(10{,}000 draws, percentile 95\% intervals, seeds kept together); a case bootstrap, exact McNemar
and a sign-flip permutation $p$ are secondary, Holm-corrected within family, in the appendix.
Cluster intervals carry no multiplicity adjustment; the Holm correction applies to the case-level
tests only. \rev{Cases cluster strongly within calculators for the library comparisons (intraclass
correlation 0.68 to 0.81), so their effective sample size is 67 to 79 cases against 120 to 190 for
the arithmetic comparisons (Table~\ref{tab:s5-paired}).}

\section{Results}
\label{sec:results}

\begin{table}[!htbp]
\centering
\small
\setlength{\tabcolsep}{3pt}
\begin{tabular}{llllcccc}
\toprule
& & & & \multicolumn{2}{c}{\textbf{Qwen2.5-7B}} & \multicolumn{2}{c}{\textbf{Qwen2.5-32B}} \\
\cmidrule(lr){5-6}\cmidrule(lr){7-8}
\textbf{Arm} & \textbf{Formula} & \textbf{Variables} & \textbf{Executes} & \shortstack{\textbf{Impl.}\\$n=440$} & \shortstack{\textbf{Full}\\$n=1{,}100$} & \shortstack{\textbf{Impl.}\\$n=440$} & \shortstack{\textbf{Full}\\$n=1{,}100$} \\
\midrule
Note only & recalled       & extracted      & model   & 25.73 & 19.62 & 33.50 & 28.67 \\
Page without formula & suppressed & \textbf{given} & model & 42.64 & 29.20 & 55.68 & 49.82 \\
Open-book arithmetic & \textbf{given} & \textbf{given} & model   & 83.64 & 72.02 & 91.45 & 83.47 \\
Extract-Solve library & hardcoded      & extracted      & function & 97.09 & 38.84 & 97.50 & 39.00 \\
Gold-Solve library & hardcoded      & \textbf{given} & function & 100.00 & 40.00 & 100.00 & 40.00 \\
\midrule
\textbf{Program-Solve} & \textbf{given} & \textbf{given} & \textbf{executor} & \rev{\textbf{84.18}} & \textbf{75.31} & \textbf{98.64} & \textbf{90.53} \\
\quad + syntax (exploratory) & \textbf{given} & \textbf{given} & executor & 87.27 & 77.95 & 98.45 & 90.96 \\
Blind Program-Solve & recalled     & extracted      & executor & 39.59 & 28.04 & 57.55 & 44.71 \\
\bottomrule
\end{tabular}
\par\smallskip
\begin{tabular}{lrrr}
\toprule
\textbf{Other checkpoints} & \textbf{Code, full} & \textbf{Arithmetic, full} & \textbf{Gap (pp)} \\
\midrule
Mistral-7B-Instruct-v0.3 & 31.44 & 39.05 & $-7.62$ \\
Phi-3.5-mini-instruct & 57.13 & 52.98 & $+4.15$ \\
\bottomrule
\end{tabular}
\caption{Accuracy (\%), five seeds. \textbf{Impl.}: the 440 cases of the 22 calculators the library
implements; \textbf{Full}: all 1{,}100. Gold-Solve is correct on all 440 by audit and abstains
elsewhere, so 40.00 by construction; both library rows abstain rather than guess. Both Open-book
arithmetic and every program arm now read the whole note (Note only and Page without formula still
cap at 120 and 250 words). Syntax lines are a benchmark-specific ablation. Dash: no formula given.}
\label{tab:ladder}
\end{table}

\subsection{Matched execution and partial-library comparisons}

Given the same formula text, gold variables and note access, Program-Solve is not reliably more
accurate than Open-book arithmetic at 7B but is at 32B, clear of zero (Table~\ref{tab:gaps}).
Program-Solve returns no valid answer on \rev{6.7\%
and 0.7\%} of case-seed rows and a wrong answer on 18.0\% and 8.8\%, against none unanswered
and 28.0\% and 16.5\% wrong for Open-book arithmetic (Table~\ref{tab:s12-breakdown}).

The library comparison is a different story from the matched one above: most of the difference
comes from coverage rather than from execution. Against the Gold-Solve library, correct on every case it implements, Program-Solve leads by
$+35.31$pp at 7B and $+50.53$pp at 32B on the full set (Table~\ref{tab:gaps}), because the library
abstains on the 660 cases it does not implement; on the 440 it does, it reaches 100\% against
84.20\% and 98.64\% for Program-Solve, and answering the other 660 replaces abstentions with some
wrong answers (Table~\ref{tab:s6-abstention}). Restricted to the 39 audit-clean calculators
(Table~\ref{tab:s10-audit-gaps}), the gap is $+32.26$ ($[15.51, 48.21]$) at 7B and $+47.59$
($[32.97, 61.79]$) at 32B. A Gold-first hybrid (library on its 440, Program-Solve elsewhere) reaches
81.64\% and 91.07\%, above every single arm; with Open-book arithmetic as fallback it reaches
78.56\% and 86.89\% ($+3.07$/$+4.18$pp for the program route, both intervals crossing zero), while
an Extract-first hybrid's program fallback is worse ($-26.44$/$-25.84$pp, both clear of zero;
Tables~\ref{tab:s13-hybrids},~\ref{tab:s18-fallbacks}).

Two syntax and date lines added to the prompt (Program-Solve + syntax) were selected on the test
split and are exploratory (Table~\ref{tab:gaps}, lower block). They lift 7B to 77.95\%,
$+2.64$pp over Program-Solve without them (cluster CI $[-0.64, 6.29]$); 32B moves little on top of
its already-clear advantage ($+0.44$pp, $[-1.38, 2.35]$). Across the four
checkpoints from three model families their effect on the Program-Solve/arithmetic gap
ranges from $-4.5$ to $+2.6$pp (Table~\ref{tab:s3-syntax}).

\begin{table}[!htbp]
\centering
\footnotesize
\setlength{\tabcolsep}{1.5pt}
\resizebox{\linewidth}{!}{%
\begin{tabular}{llrr}
\toprule
\textbf{Comparison} & \textbf{Access} & \textbf{7B} & \textbf{32B} \\
\midrule
Program-Solve $-$ Open-book arithmetic & formula, gold variables, whole note & $+3.29$ $[-3.49, 10.38]$ & $+7.05$ $[0.47, 14.60]$ \\
Program-Solve $-$ Gold-Solve library & gold variables & $+35.31$ $[21.87, 48.11]$ & $+50.53$ $[38.00, 62.80]$ \\
Blind Program-Solve $-$ Extract-Solve library & both extract & $-10.80$ $[-24.35, 2.64]$ & $+5.71$ $[-7.96, 18.87]$ \\
\midrule
\multicolumn{4}{l}{\emph{Exploratory, prompt selected on the test split}} \\
Program-Solve + syntax $-$ Open-book arithmetic & formula, gold variables, whole note & $+5.93$ $[-0.11, 12.55]$ & $+7.49$ $[1.00, 15.16]$ \\
Program-Solve + syntax $-$ Gold-Solve library & gold variables & $+37.95$ $[25.24, 50.38]$ & $+50.96$ $[38.47, 63.09]$ \\
\bottomrule
\end{tabular}%
}
\caption{Paired gaps (pp) on the full 1{,}100, five seeds, with 95\% calculator-cluster bootstrap
intervals (10{,}000 draws), unadjusted for multiplicity. Upper block: the original prompt. Lower
block: the syntax-added prompt, selected on the test split. Intervals crossing zero establish
neither difference nor equivalence; paired estimates can differ slightly from rounded-mean
differences.}
\label{tab:gaps}
\end{table}

\subsection{Removing formula and gold-variable access}

Blind Program-Solve reaches 28.04\% at 7B and 44.71\% at 32B, declines of 47.27 and
45.82pp from Program-Solve (Table~\ref{tab:ladder}); formula and variable access change
together, so the design does not separate recall from extraction. Against the Extract-Solve
library the point estimate is lower at 7B and higher at 32B, but both intervals cross
zero (Table~\ref{tab:gaps}). \rev{The observed failures are formula and variable errors, read from outputs without a controlled decomposition:} an
ideal-body-weight convention in Cockcroft-Gault, potassium in a corrected anion gap, heart rate for
respiratory rate.

\paragraph{Other families.} Table~\ref{tab:ladder} includes the Mistral and Phi results,
full set, formula-given code against arithmetic, both reading the whole note: secondary
permutation $p<0.001$ and $p=0.010$; the two move in opposite directions, Mistral's
code route 7.6pp behind arithmetic and Phi-3.5's 4.2pp ahead. Mistral's Extract-Solve library
reaches 34.89\%, above either Program-Solve arm.

\section{Limitations}
\label{sec:limitations}

\paragraph{Experimental scope.} Two Qwen checkpoints do not establish scaling; Mistral and Phi-3.5
move in opposite directions from each other and from both Qwen checkpoints. Four delta-gap calculators received the plain anion-gap formula until the
program audit found it, and two more (an anion-gap variant and a related osmolality
calculator) were found aliased onto the wrong quantity in a second pass; formula-reading arms
were rerun after each fix. A completeness audit of all 55
supplied texts (Table~\ref{tab:s17-catalogue}) then found 28 wrong or incomplete, 10 unable to
reproduce the benchmark's number; every arm in this rerun, including Open-book arithmetic and
Program-Solve, read the same, fully corrected texts. \rev{Re-auditing the corrected runs (Table~\ref{tab:s15-failure-audit}), no sampled error traces to an under-specified formula; the residual failures are unit conversions invented for supplied inputs and one-tier slips inside multi-band scoring tables, so Program-Solve fails on arithmetic hygiene rather than on clinical knowledge.} Note only reads 120 words, understating a budget-matched
baseline by 4 to 12 points, and the comparator is a partial 22-calculator library. The residual
failures are the errors a tired clinician makes, invented unit conversions and one-tier slips in
banded scores, and the errors a validated calculator never makes; the library's 100\% on its 440
covered cases shows what abstention is worth.

\paragraph{Clinical validity.} Every gold answer is the calculator's own output, so final-answer
accuracy validates neither the program's logic nor the formula. A preliminary literature-based
audit of all 55 calculators (Tables~\ref{tab:s9-audit} to \ref{tab:s11-audit-tiers}) flags 16 with
a version, use or coefficient concern, four of them replaced by a current guideline, which the
benchmark still rewards reproducing exactly. On those four, Program-Solve scores 71.00 and 90.00
against 28.25 and 36.50 for Open-book arithmetic, its largest gap over arithmetic at both scales;
removing them moves no headline gap outside its interval
(Table~\ref{tab:s10-audit-gaps}). The larger point is that formula provenance and version should be
explicit inputs to any calculator interface, human or automated, rather than assumptions inherited
from the benchmark.

\paragraph{Deployment.} No Global South data, language or locale is evaluated; the benchmark is
English with US conventions, and the cost of local serving is not measured. Notes in other
languages, laboratory values in mmol/L rather than mg/dL and day/month/year dates each open a
further path to the unit-conversion failures observed here; the exploratory date-format prompt
lines show how much locale the current result silently assumes.

\section{Conclusion}

With the formula, variables and note access matched, an open-weight model writing a program is
not reliably more accurate than the same model doing the arithmetic at 7B scale, but is at 32B
scale (+7.1pp, cluster CI clear of zero); its advantage over a partial hand-written library at
either scale comes mostly from answering where the library abstains; \rev{where both answer, the
library is the more accurate}. Which of these two patterns a given open-weight checkpoint will show is not yet
predictable from scale alone: Mistral-7B and Phi-3.5-mini move in opposite directions on the
same comparison. Pending that answer, the clinically defensible configuration is a verified
library where one exists, program generation where it does not, and explicit abstention where
neither can be trusted. The next question is what separates them. Code and data:
\url{https://github.com/felipeocampoos/Towards-a-Deterministic-Math-Solver-for-Clinical-Language-Models}.

\section*{Acknowledgments}
This research was supported by Anthropic's AI for Science program. GPU compute was
provided by NVIDIA through the Brev academic grant node and by the MIT Office of Research
Computing and Data (ORCD) cluster.

\bibliographystyle{unsrt}
{\bibsize\bibliography{\bibpath}}

\clearpage
\appendix
\section{Appendix}
\label{sec:appendix}

\subsection{Arm protocols and settings}
\label{sec:protocols}

Every arm calls the served model through one chat request per turn with only temperature and the
output-token limit set; Table~\ref{tab:s14-settings} lists every setting and executor limit, and
the repository linked in the Conclusion holds the implementation of every arm. All
single-call arms decode greedily (temperature 0); sampled candidates use temperature 0.7. Every
arm runs five seeds (42 to 46); the vote and repair levers run on the covered 440 only. The seed
sets the case order and the run identity; it never reaches the request, so the main arms are five
repeated greedy runs. The worked example is a fixed benchmark case except in the note-only arms,
where the seed chooses it. Repeated runs still differ through server batching: agreement across the
five seeds is 78.5 to 100\% for the greedy arms and 55.6 to 75.2\% where the example moves
(Table~\ref{tab:s19-seeds}).
The benchmark's test split circulates in three copies that disagree on a few dozen rows; every run
read the Hugging Face parquet at revision 5488179, \rev{which the runner pins and downloads at start; the file
itself is not redistributed with the code, and every number here is scored against that revision}.

\paragraph{Note only.} One call: a step-by-step instruction, one worked example from another
benchmark case (its note cut to 24 words, its explanation to 40 words, and its answer; the example
is chosen by the seed), then the case with the first 120 words of its note. Note only with gold
variables (the row labelled Note only with gold variables in Table~\ref{tab:s1-ladders}) adds the gold
variable block to the same single call; the code has no second pass.

\paragraph{Page without formula and Open-book arithmetic.} One greedy call with a zero-shot
chain-of-thought instruction, the gold variable block and the question; Page without formula reads the first 250 words of the note and omits the calculator's formula text, Open-book
arithmetic reads the whole note and adds the formula text.

\paragraph{Extract-Solve and Gold-Solve library.} Extract-Solve makes one greedy extraction call
over the whole note, naming the calculator and listing the variables its hand-written function
needs, then runs that function; it abstains outside the 22 implemented calculators or when the
extraction is invalid. Gold-Solve gives the same functions the gold variables. Extract-Solve
library, recalled (Table~\ref{tab:s1-ladders}) first samples three chain-of-thought answers at
temperature 0.7 from the Note only prompt (one fixed worked example); if at least two agree and
none is empty, that answer is returned. Otherwise, when the calculator is one of the 22, one greedy extraction call over the
whole note feeds the hand-written function; when it is not, the majority answer, or the first
sample when there is no majority, is returned, so all arithmetic outside the library is the
model's. It abstains only when a routed extraction is invalid or the function raises (30 of
5{,}500 rows at 7B, 45 at 32B) and costs three model calls per case, four when it routes (35.1\%
of rows at 7B, 29.7\% at 32B).

\paragraph{Formulate-Solve tree.} One greedy call over the whole note, with no variables,
calculator name or formula, asking for a structured reply with the calculation name, a one-line formula, the
numeric variables, a list of missing variables and an expression tree over basic arithmetic
operators and comparisons; the tree is evaluated exactly. It abstains when the tree or the
variables are invalid, when a referenced variable is listed as missing, or when evaluation raises;
the 60 date cases abstain before any call.

\paragraph{Program-Solve and Blind Program-Solve.} One greedy program call over the whole note,
run by the executor at an output-token budget of 2{,}048 for Program-Solve and its syntax variant
and 1{,}024 for Blind Program-Solve and its vote/repair variants \rev{(across the formula-reading arms, at
most 2.1\% of calls in any family end at the output limit, 0.2\% or fewer at 32B, and no truncated
call scores correct)}; the Program-Solve prompt carries the formula text and
the gold variables, the Blind Program-Solve prompt neither. The + syntax variants insert two lines
before the instruction to return one code block, both about the language and the note, not the
calculation: variable names must be valid Python identifiers (lowercase words joined by
underscores, never the calculation's name), and the notes write dates as month/day/year, so the
program must build each date from the three numbers itself and add or subtract days with the
standard date library. Blind + vote samples
five programs at temperature 0.7, runs each, and returns the value that occurs at least twice
(ties go to the value sampled first), else the first computed answer; it abstains if none runs.
Blind + repair makes one greedy program call and, if the executor returns no value, one corrective
turn that shows the original prompt, the previous reply and the executor's own error string, never
anything about the calculation; one or two calls per case. Blind + repair + vote combines both,
five to eight calls per case.

\paragraph{Scoring.} A numeric answer is the first number on the reply's final-answer line and is
correct when it lies in the benchmark's own interval: gold plus or minus 5\% for decimal outputs
and exactly the gold for integers; a reply with no parseable number scores wrong. Date answers are
parsed to a calendar day and must match the gold day; gestational ages are reduced to their
(weeks, days) integers and must match exactly.

\paragraph{Covered-subset runs.} Blind Program-Solve on the 440 covered cases is 39.59 / 57.55 in
Table~\ref{tab:ladder}, the covered subset of the full-set run, and 39.73 / 57.59 in
Table~\ref{tab:s2-levers}, a separate run restricted to those cases with the same prompt; the
per-case answers agree on about 95\% (7B) and at least 99.8\% (32B) of case-seed pairs and the
difference is re-run noise.

\subsection{Supplementary tables}

Tables~\ref{tab:s1-ladders} to \ref{tab:s8-worked} and \ref{tab:s12-breakdown} to
\ref{tab:s14-settings} are produced from the committed per-case results by the analysis scripts in
the repository; Tables~\ref{tab:s9-audit} to \ref{tab:s11-audit-tiers} come from the calculator
audit in the same repository.

\begin{itemize}\itemsep0pt
  \item Table~\ref{tab:s1-ladders}: full ladders, four families, every arm that ran.
  \item Table~\ref{tab:s2-levers}: blind Program-Solve levers, vote and repair, on the covered 440.
  \item Table~\ref{tab:s3-syntax}: the syntax-lines ablation.
  \item Table~\ref{tab:s4-notebudget}: the 250-word note-budget ablation.
  \item Table~\ref{tab:s5-paired}: every paired gap, with case and calculator-cluster intervals.
  \item Table~\ref{tab:s6-abstention}: abstention accounting for the program arms.
  \item Table~\ref{tab:s7-recall}: calculators the blind arm never gets right, with the formula text and the gold variables both withheld.
  \item Table~\ref{tab:s8-worked}: one case, formula given against formula recalled.
  \item Table~\ref{tab:s9-audit}: the 16 calculators with a concern, four of them replaced by a guideline, by concern type, with sources.
  \item Table~\ref{tab:s10-audit-gaps}: the four headline gaps recomputed on the not-replaced and no-issue-identified calculator sets.
  \item Table~\ref{tab:s11-audit-tiers}: accuracy by audit group.
  \item Table~\ref{tab:s12-breakdown}: right, wrong and no-answer rates on the supported 440, the unsupported 660 and the full 1{,}100.
  \item Table~\ref{tab:s13-hybrids}: library-first hybrid baselines with paired intervals.
  \item Table~\ref{tab:s14-settings}: reproducibility ledger, decoding, note budgets and executor limits.
  \item Table~\ref{tab:s16-entitynone}: note budget against variable access.
  \item Table~\ref{tab:s17-catalogue}: completeness audit of the 55 supplied formula texts.
  \item Table~\ref{tab:s18-fallbacks}: library-first baselines with each fallback route.
  \item Table~\ref{tab:s19-seeds}: agreement across the five seeds, by arm.
\end{itemize}

\begin{figure}[!htbp]
\centering
\includegraphics[width=0.62\linewidth]{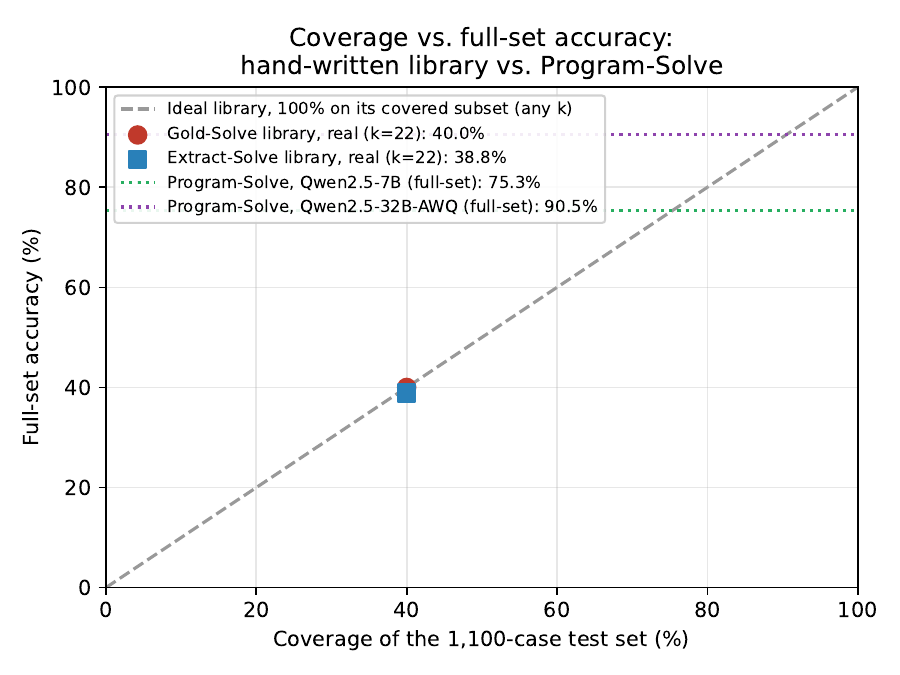}
\caption{\rev{Coverage against full-set accuracy for a library that abstains outside the calculators it
implements. The benchmark is balanced, so a library of $k$ calculators covers $20k$ of 1{,}100 cases
and its full-set accuracy cannot exceed that share; the 22-calculator library sits at 40\%.
Program-Solve supports all 55 calculators and attempts every case, though it does not return a
valid answer on every one (Table~\ref{tab:s6-abstention}).}}
\label{fig:coverage}
\end{figure}

% Float pages stack their tables from the top; the slack goes below the last one.
\makeatletter
\setlength{\@fptop}{0pt}
\setlength{\@fpsep}{10pt plus 2pt}
\setlength{\@fpbot}{0pt plus 1fil}
\makeatother

\begin{table}[!htbp]
\centering
\footnotesize
\setlength{\tabcolsep}{4pt}
\begin{tabular*}{\linewidth}{@{\extracolsep{\fill}}lrrrr@{}}
\toprule
\textbf{Arm} & \textbf{Qwen2.5-7B} & \textbf{Qwen2.5-32B-AWQ} & \textbf{Mistral-7B-v0.3} & \textbf{Phi-3.5-mini} \\
\midrule
Note only & 19.6\,{\tiny 1.3} & 28.7\,{\tiny 1.6} & 13.3\,{\tiny 0.9} & 18.1\,{\tiny 1.0} \\
Note only with gold variables & 31.4\,{\tiny 1.8} & 48.4\,{\tiny 1.8} & 18.9\,{\tiny 1.1} & 27.1\,{\tiny 1.5} \\
Open-book arithmetic & 72.0\,{\tiny 0.4} & 83.5\,{\tiny 0.2} & 39.1\,{\tiny 0.2} & 53.0\,{\tiny 0.6} \\
Page without formula & 29.2\,{\tiny 0.5} & 49.8\,{\tiny 0.2} & 21.2\,{\tiny 0.7} & 25.9\,{\tiny 0.4} \\
Extract-Solve library & 38.8\,{\tiny 0.1} & 39.0\,{\tiny 0.0} & 34.9\,{\tiny 0.1} & 35.0\,{\tiny 0.4} \\
Extract-Solve library, recalled & 46.9\,{\tiny 0.2} & 50.3\,{\tiny 1.2} & 40.4\,{\tiny 0.4} & 41.1\,{\tiny 1.0} \\
Formulate-Solve tree & 17.0\,{\tiny 0.4} & 23.8\,{\tiny 0.1} & 4.6\,{\tiny 0.1} & 0.5\,{\tiny 0.1} \\
Blind Program-Solve & 28.0\,{\tiny 0.5} & 44.7\,{\tiny 0.2} & 11.2\,{\tiny 0.3} & 23.4\,{\tiny 0.6} \\
Program-Solve & 75.3\,{\tiny 0.5} & 90.5\,{\tiny 0.1} & 31.4\,{\tiny 0.8} & 57.1\,{\tiny 0.8} \\
Program-Solve + syntax & 77.9\,{\tiny 0.4} & 91.0\,{\tiny 0.1} & 26.9\,{\tiny 0.7} & 57.3\,{\tiny 0.3} \\
Blind + vote & 43.5\,{\tiny 1.1}$^{440}$ & 57.7\,{\tiny 0.7}$^{440}$ & 21.4\,{\tiny 1.3}$^{440}$ & 35.4\,{\tiny 1.3}$^{440}$ \\
Blind + repair & 39.8\,{\tiny 0.5}$^{440}$ & 57.5\,{\tiny 0.1}$^{440}$ & 14.8\,{\tiny 0.3}$^{440}$ & 31.1\,{\tiny 0.8}$^{440}$ \\
Blind + repair + vote & 43.4\,{\tiny 1.1}$^{440}$ & 57.7\,{\tiny 0.7}$^{440}$ & 21.0\,{\tiny 1.1}$^{440}$ & 37.5\,{\tiny 1.4}$^{440}$ \\
Extract-Solve library, 250 words & 32.3\,{\tiny 0.1} & 32.3\,{\tiny 0.0} & 29.3\,{\tiny 0.1} & 28.2\,{\tiny 0.3} \\
Extract-Solve library, recalled, 250 words & 39.9\,{\tiny 0.5} & 45.1\,{\tiny 0.6} & 36.6\,{\tiny 0.6} & 36.9\,{\tiny 1.2} \\
\bottomrule
\end{tabular*}
\caption{Accuracy in percent on the MedCalc-Bench test split: mean over seeds, seed standard deviation in small type. Unmarked cells are the full 1,100 cases; $^{440}$ marks the covered subset, $^{1}$ a single seed.}
\label{tab:s1-ladders}
\end{table}

\begin{table}[!htbp]
\centering
\footnotesize
\setlength{\tabcolsep}{4pt}
\begin{tabular*}{\linewidth}{@{\extracolsep{\fill}}lrrrr@{}}
\toprule
\textbf{Arm} & \textbf{Qwen2.5-7B} & \textbf{Qwen2.5-32B-AWQ} & \textbf{Mistral-7B-v0.3} & \textbf{Phi-3.5-mini} \\
\midrule
Blind Program-Solve & 39.7\,{\tiny 0.7} & 57.6\,{\tiny 0.1} & 12.6\,{\tiny 0.3} & 29.9\,{\tiny 1.0} \\
+ vote & 43.5\,{\tiny 1.1} & 57.7\,{\tiny 0.7} & 21.4\,{\tiny 1.3} & 35.4\,{\tiny 1.3} \\
+ repair & 39.8\,{\tiny 0.5} & 57.5\,{\tiny 0.1} & 14.8\,{\tiny 0.3} & 31.1\,{\tiny 0.8} \\
+ repair + vote & 43.4\,{\tiny 1.1} & 57.7\,{\tiny 0.7} & 21.0\,{\tiny 1.1} & 37.5\,{\tiny 1.4} \\
\bottomrule
\end{tabular*}
\caption{Blind Program-Solve levers on the covered 440, where every family has data. Voting is over five sampled programs; the repair turn shows the executor error back to the model once. Mean over seeds, seed standard deviation in small type.}
\label{tab:s2-levers}
\end{table}

\begin{table}[!htbp]
\centering
\footnotesize
\setlength{\tabcolsep}{4pt}
\begin{tabular*}{\linewidth}{@{\extracolsep{\fill}}lrrrr@{}}
\toprule
\textbf{Arm} & \textbf{Qwen2.5-7B} & \textbf{Qwen2.5-32B-AWQ} & \textbf{Mistral-7B-v0.3} & \textbf{Phi-3.5-mini} \\
\midrule
Program-Solve & 75.3\,{\tiny 0.5} & 90.5\,{\tiny 0.1} & 31.4\,{\tiny 0.8} & 57.1\,{\tiny 0.8} \\
Program-Solve + syntax & 77.9\,{\tiny 0.4} & 91.0\,{\tiny 0.1} & 26.9\,{\tiny 0.7} & 57.3\,{\tiny 0.3} \\
\bottomrule
\end{tabular*}
\caption{The two generic Python lines name no calculator, formula or clinical quantity: variable names must be valid identifiers, and the notes write dates as month/day/year. Each column pairs the two arms on the same subset, the full 1,100 where both exist, else the covered 440 ($^{440}$).}
\label{tab:s3-syntax}
\end{table}

\begin{table}[!htbp]
\centering
\footnotesize
\setlength{\tabcolsep}{4pt}
\begin{tabular*}{\linewidth}{@{\extracolsep{\fill}}lrrrr@{}}
\toprule
\textbf{Arm} & \textbf{Qwen2.5-7B} & \textbf{Qwen2.5-32B-AWQ} & \textbf{Mistral-7B-v0.3} & \textbf{Phi-3.5-mini} \\
\midrule
Extract-Solve library, uncapped & 38.8\,{\tiny 0.1} & 39.0\,{\tiny 0.0} & 34.9\,{\tiny 0.1} & 35.0\,{\tiny 0.4} \\
Extract-Solve library, 250 words & 32.3\,{\tiny 0.1} & 32.3\,{\tiny 0.0} & 29.3\,{\tiny 0.1} & 28.2\,{\tiny 0.3} \\
Extract-Solve library, recalled, uncapped & 46.9\,{\tiny 0.2} & 50.3\,{\tiny 1.2} & 40.4\,{\tiny 0.4} & 41.1\,{\tiny 1.0} \\
Extract-Solve library, recalled, 250 words & 39.9\,{\tiny 0.5} & 45.1\,{\tiny 0.6} & 36.6\,{\tiny 0.6} & 36.9\,{\tiny 1.2} \\
\bottomrule
\end{tabular*}
\caption{Note-budget ablation on the full 1,100. The extracting arms read the whole note by default, as the program arms and Open-book arithmetic do; the ablation caps them at the 250 words Page without formula reads. The one-shot note arms read 120. Table~\ref{tab:s16-entitynone} moves note budget and variable access separately.}
\label{tab:s4-notebudget}
\end{table}

\begin{table}[!htbp]
\centering
\footnotesize
\setlength{\tabcolsep}{4pt}
\begin{tabular*}{\linewidth}{@{\extracolsep{\fill}}lrrrrrrrr@{}}
\toprule
\textbf{Gap} & \textbf{$n$} & \textbf{pp} & \textbf{Perm.\ $p$} & \textbf{Holm $p$} & \textbf{Case CI} & \textbf{Cluster CI} & \textbf{ICC} & \textbf{$n_{\mathrm{eff}}$} \\
\midrule
\multicolumn{9}{@{}l}{\emph{Qwen2.5-7B}} \\
Prog.+syntax $-$ Arith. & 1,100 & +5.93 & $<$0.001 & $<$0.001 & $[3.16,\,8.69]$ & $[-0.11,\,12.55]$ & 0.23 & 206 \\
Prog.+syntax $-$ Gold lib. & 1,100 & +37.95 & $<$0.001 & $<$0.001 & $[34.56,\,41.24]$ & $[25.24,\,50.38]$ & 0.65 & 82 \\
Prog. $-$ Gold lib. & 1,100 & +35.31 & $<$0.001 & $<$0.001 & $[31.80,\,38.73]$ & $[21.87,\,48.11]$ & 0.68 & 79 \\
Prog. $-$ Arith. & 1,100 & +3.29 & 0.021 & 0.033 & $[0.47,\,6.13]$ & $[-3.49,\,10.38]$ & 0.25 & 190 \\
Prog.+syntax $-$ Prog. & 1,100 & +2.64 & 0.016 & 0.033 & $[0.45,\,4.80]$ & $[-0.64,\,6.29]$ & 0.10 & 390 \\
Blind $-$ Arith. & 1,100 & -43.98 & $<$0.001 & $<$0.001 & $[-46.95,\,-40.93]$ & $[-53.40,\,-34.49]$ & 0.41 & 124 \\
Blind $-$ Extract lib. & 1,100 & -10.80 & $<$0.001 & $<$0.001 & $[-14.18,\,-7.38]$ & $[-24.35,\,2.64]$ & 0.73 & 74 \\
Blind $-$ Recalled lib. & 1,100 & -18.84 & $<$0.001 & $<$0.001 & $[-21.85,\,-15.84]$ & $[-29.91,\,-8.04]$ & 0.51 & 103 \\
Blind+vote $-$ Blind & 440 & +3.82 & $<$0.001 & 0.002 & $[1.64,\,6.14]$ & $[0.95,\,7.05]$ & 0.05 & 229 \\
\addlinespace
\multicolumn{9}{@{}l}{\emph{Qwen2.5-32B-AWQ}} \\
Prog.+syntax $-$ Gold lib. & 1,100 & +50.96 & $<$0.001 & $<$0.001 & $[47.95,\,53.98]$ & $[38.47,\,63.09]$ & 0.81 & 67 \\
Prog. $-$ Gold lib. & 1,100 & +50.53 & $<$0.001 & $<$0.001 & $[47.45,\,53.55]$ & $[38.00,\,62.80]$ & 0.81 & 67 \\
Prog.+syntax $-$ Arith. & 1,100 & +7.49 & $<$0.001 & $<$0.001 & $[5.07,\,9.98]$ & $[1.00,\,15.16]$ & 0.40 & 128 \\
Prog. $-$ Arith. & 1,100 & +7.05 & $<$0.001 & $<$0.001 & $[4.71,\,9.47]$ & $[0.47,\,14.60]$ & 0.43 & 120 \\
Prog.+syntax $-$ Prog. & 1,100 & +0.44 & 0.525 & 0.525 & $[-0.84,\,1.71]$ & $[-1.38,\,2.35]$ & 0.10 & 390 \\
Blind $-$ Arith. & 1,100 & -38.76 & $<$0.001 & $<$0.001 & $[-41.96,\,-35.56]$ & $[-48.40,\,-29.40]$ & 0.43 & 121 \\
Blind $-$ Extract lib. & 1,100 & +5.71 & 0.002 & 0.004 & $[2.11,\,9.29]$ & $[-7.96,\,18.87]$ & 0.69 & 78 \\
Blind $-$ Recalled lib. & 1,100 & -5.56 & 0.001 & 0.003 & $[-8.91,\,-2.24]$ & $[-16.33,\,4.64]$ & 0.42 & 123 \\
\addlinespace
\multicolumn{9}{@{}l}{\emph{Mistral-7B-v0.3}} \\
Prog. $-$ Arith. & 1,100 & -7.62 & $<$0.001 & $<$0.001 & $[-11.20,\,-4.05]$ & $[-16.42,\,1.40]$ & 0.27 & 181 \\
Prog. $-$ Arith. & 440 & -4.86 & 0.113 & 0.113 & $[-10.91,\,1.05]$ & $[-19.77,\,11.09]$ & 0.30 & 65 \\
Prog. $-$ Extract lib. & 1,100 & -3.45 & 0.052 & 0.104 & $[-7.00,\,0.04]$ & $[-15.36,\,8.16]$ & 0.50 & 105 \\
Blind+vote $-$ Blind & 440 & +8.73 & $<$0.001 & $<$0.001 & $[6.09,\,11.41]$ & $[3.36,\,14.77]$ & 0.11 & 144 \\
\addlinespace
\multicolumn{9}{@{}l}{\emph{Phi-3.5-mini}} \\
Prog. $-$ Arith. & 1,100 & +4.15 & 0.010 & 0.010 & $[0.96,\,7.24]$ & $[-3.56,\,12.22]$ & 0.27 & 181 \\
Prog. $-$ Arith. & 440 & -25.05 & $<$0.001 & $<$0.001 & $[-30.14,\,-20.00]$ & $[-40.09,\,-8.27]$ & 0.41 & 51 \\
\bottomrule
\end{tabular*}
\caption{Every paired gap, solver minus reference. Prog.\ is Program-Solve, Arith.\ Open-book arithmetic, Gold lib., Extract lib.\ and Recalled lib.\ the Gold-Solve, Extract-Solve and recalled Extract-Solve libraries, Blind the blind program arm. Permutation $p$ is a sign-flip test on seed-averaged per-case differences, Holm-adjusted within family; the cluster interval resamples calculators.}
\label{tab:s5-paired}
\end{table}

\begin{table}[!htbp]
\centering
\scriptsize
\setlength{\tabcolsep}{3pt}
\begin{tabular*}{\linewidth}{@{\extracolsep{\fill}}llrrrrrrrrrr@{}}
\toprule
\multicolumn{3}{c}{} & \multicolumn{5}{c}{\textbf{No valid answer, rows}} & \multicolumn{1}{c}{} & \multicolumn{3}{c}{\textbf{Accuracy, \%}} \\
\cmidrule(lr){4-8} \cmidrule(lr){10-12}
\textbf{Arm} & \textbf{Set} & \textbf{Seeds} & \textbf{No prog.} & \textbf{Rejected} & \textbf{Raised} & \textbf{Bad kind} & \textbf{Unread.} & \textbf{None \%} & \textbf{All} & \textbf{Att.} & \textbf{Dates} \\
\midrule
\multicolumn{12}{@{}l}{\emph{Qwen2.5-7B}} \\
Blind Program-Solve & Full 1,100 & 5 & 384 & 111 & 59 & 121 & 68 & 13.5 & 28.04 & 31.96 & 44.33 \\
Program-Solve & Full 1,100 & 5 & 70 & 47 & 245 & 0 & 5 & 6.7 & 75.31 & 80.62 & 33.00 \\
Program-Solve + syntax & Full 1,100 & 5 & 0 & 110 & 128 & 0 & 2 & 4.4 & 77.95 & 81.47 & 59.00 \\
Blind + vote & Covered 440 & 5 & 0 & 0 & 0 & 1 & 11 & 0.6 & 43.55 & 43.57 & 50.00 \\
Blind + repair & Covered 440 & 5 & 0 & 0 & 13 & 97 & 60 & 7.7 & 39.77 & 41.87 & 44.67 \\
Blind + repair + vote & Covered 440 & 5 & 0 & 0 & 0 & 0 & 7 & 0.3 & 43.41 & 43.41 & 48.67 \\
\addlinespace
\multicolumn{12}{@{}l}{\emph{Qwen2.5-32B-AWQ}} \\
Blind Program-Solve & Full 1,100 & 5 & 36 & 60 & 29 & 0 & 21 & 2.6 & 44.71 & 45.75 & 71.67 \\
Program-Solve & Full 1,100 & 5 & 1 & 5 & 30 & 0 & 0 & 0.7 & 90.53 & 91.12 & 93.33 \\
Program-Solve + syntax & Full 1,100 & 5 & 0 & 8 & 29 & 0 & 0 & 0.7 & 90.96 & 91.58 & 100.00 \\
Blind + vote & Covered 440 & 5 & 0 & 0 & 1 & 0 & 1 & 0.1 & 57.73 & 57.75 & 75.33 \\
Blind + repair & Covered 440 & 5 & 0 & 0 & 5 & 0 & 5 & 0.5 & 57.45 & 57.59 & 71.67 \\
Blind + repair + vote & Covered 440 & 5 & 0 & 0 & 0 & 0 & 1 & 0.1 & 57.68 & 57.68 & 73.67 \\
\addlinespace
\multicolumn{12}{@{}l}{\emph{Mistral-7B-v0.3}} \\
Blind Program-Solve & Full 1,100 & 5 & 147 & 345 & 1159 & 492 & 18 & 39.3 & 11.16 & 18.29 & 23.67 \\
Program-Solve & Full 1,100 & 5 & 96 & 151 & 1282 & 74 & 5 & 31.8 & 31.44 & 46.06 & 42.33 \\
Program-Solve + syntax & Full 1,100 & 5 & 24 & 212 & 2071 & 111 & 12 & 45.8 & 26.95 & 49.53 & 23.33 \\
Blind + vote & Covered 440 & 5 & 0 & 3 & 20 & 10 & 14 & 2.1 & 21.36 & 21.69 & 29.00 \\
Blind + repair & Covered 440 & 5 & 3 & 49 & 216 & 129 & 7 & 18.4 & 14.77 & 18.03 & 23.00 \\
Blind + repair + vote & Covered 440 & 5 & 0 & 0 & 3 & 2 & 14 & 0.9 & 20.95 & 21.00 & 26.67 \\
\addlinespace
\multicolumn{12}{@{}l}{\emph{Phi-3.5-mini}} \\
Blind Program-Solve & Full 1,100 & 5 & 83 & 158 & 530 & 135 & 85 & 18.0 & 23.36 & 27.97 & 25.33 \\
Program-Solve & Full 1,100 & 5 & 41 & 278 & 254 & 142 & 128 & 15.9 & 57.13 & 66.12 & 34.00 \\
Program-Solve + syntax & Full 1,100 & 5 & 78 & 212 & 281 & 130 & 156 & 15.6 & 57.25 & 65.63 & 42.33 \\
Blind + vote & Covered 440 & 5 & 0 & 1 & 23 & 3 & 29 & 2.5 & 35.41 & 35.85 & 37.00 \\
Blind + repair & Covered 440 & 5 & 0 & 15 & 124 & 41 & 30 & 9.6 & 31.14 & 33.91 & 30.00 \\
Blind + repair + vote & Covered 440 & 5 & 0 & 0 & 1 & 1 & 41 & 1.9 & 37.45 & 37.49 & 43.67 \\
\bottomrule
\end{tabular*}
\caption{Every case-seed row of the program arms under the outcome taxonomy shared with Table~\ref{tab:s12-breakdown}: no program written, sandbox rejection, raised error, wrong return kind, or a value the scorer could not read. Counts pool the seeds shown. Att.\ is accuracy on attempted rows; Dates, on the 60 date cases.}
\label{tab:s6-abstention}
\end{table}

\begin{table}[!htbp]
\centering
\footnotesize
\setlength{\tabcolsep}{4pt}
\begin{tabular*}{\linewidth}{@{\extracolsep{\fill}}lrrr@{}}
\toprule
\textbf{Calculator} & \textbf{Program right} & \textbf{Blind wrong} & \textbf{\%} \\
\midrule
\multicolumn{4}{@{}l}{\emph{Qwen2.5-7B}} \\
Delta Gap & 100 & 100 & 100 \\
Maintenance Fluids Calculations & 100 & 100 & 100 \\
Steroid Conversion Calculator & 100 & 95 & 95 \\
Adjusted Body Weight & 100 & 94 & 94 \\
QTc Rautaharju Calculator & 100 & 90 & 90 \\
QTc Fridericia Calculator & 100 & 99 & 99 \\
\addlinespace
\multicolumn{4}{@{}l}{\emph{Qwen2.5-32B-AWQ}} \\
MDRD GFR Equation & 100 & 90 & 90 \\
Delta Gap & 100 & 95 & 95 \\
Albumin Corrected Anion Gap & 100 & 90 & 90 \\
CKD-EPI Equations for Glomerular Filtration Rate & 100 & 90 & 90 \\
Albumin Corrected Delta Ratio & 100 & 100 & 100 \\
Delta Ratio & 100 & 100 & 100 \\
\addlinespace
\multicolumn{4}{@{}l}{\emph{Mistral-7B-v0.3}} \\
Maintenance Fluids Calculations & 96 & 96 & 100 \\
QTc Fridericia Calculator & 82 & 79 & 96 \\
QTc Bazett Calculator & 76 & 75 & 99 \\
Estimated Date of Conception & 44 & 44 & 100 \\
QTc Framingham Calculator & 43 & 43 & 100 \\
Framingham Risk Score for Hard Coronary Heart Disease & 42 & 42 & 100 \\
\addlinespace
\multicolumn{4}{@{}l}{\emph{Phi-3.5-mini}} \\
QTc Framingham Calculator & 96 & 90 & 94 \\
Delta Ratio & 94 & 94 & 100 \\
Delta Gap & 90 & 88 & 98 \\
Free Water Deficit & 74 & 73 & 99 \\
Adjusted Body Weight & 68 & 67 & 98 \\
Morphine Milligram Equivalents (MME) Calculator & 64 & 59 & 92 \\
\bottomrule
\end{tabular*}
\caption{Calculators the blind arm loses on every seed once the supplied formula and the gold variable list are both removed, execution held fixed: case-seed pairs, five seeds, where the formula-given program is correct and the blind one is not. At least ten such pairs, top 6 per family.}
\label{tab:s7-recall}
\end{table}

\begin{table}[!htbp]
\centering
\footnotesize
\setlength{\tabcolsep}{4pt}
\begin{tabular*}{\linewidth}{@{\extracolsep{\fill}}p{2.4cm}p{7.4cm}rl@{}}
\toprule
{\raggedright \textbf{Condition}\par} & {\raggedright \textbf{How the program set the weight}\par} & \textbf{Answer} & \textbf{Scored} \\
\midrule
{\raggedright Formula, variables and syntax lines given\par} & {\raggedright Derives an ideal body weight of 52.8\,kg from the height, finds the BMI of 25.2 is not below 25, and so puts the smaller of actual and ideal weight, 52.8\,kg, into the Cockcroft-Gault equation with the female factor.\par} & 40.96 & correct \\
{\raggedright None of the three given\par} & {\raggedright Derives an ideal body weight of 47.8\,kg from the height in inches, finds the BMI of 25.2 is above 25, and on that branch keeps the actual weight, 61.0\,kg, in the Cockcroft-Gault equation with the female factor.\par} & 46.81 & wrong \\
\bottomrule
\end{tabular*}
\caption{One case (Creatinine Clearance (Cockcroft-Gault Equation), seed 42, Qwen2.5-7B), gold 40.97. The same model writes both programs and both implement Cockcroft-Gault. The upper row is given the calculator name, its formula, the gold variable list and two generic Python lines; the lower row none of them.}
\label{tab:s8-worked}
\end{table}

\begin{table}[!htbp]
\centering
\scriptsize
\setlength{\tabcolsep}{3pt}
\renewcommand{\arraystretch}{0.92}
\begin{tabular*}{\linewidth}{@{\extracolsep{\fill}}p{2.1cm}p{2.9cm}p{1.8cm}p{6.5cm}@{}}
\toprule
{\raggedright \textbf{Calculator}\par} & {\raggedright \textbf{Benchmark formula}\par} & {\raggedright \textbf{Type; use}\par} & {\raggedright \textbf{Body, jurisdiction, date; concern; sources}\par} \\
\midrule
{\raggedright Framingham Risk Score for Hard Coronary Heart Disease\par} & {\raggedright Framingham hard CHD 10-year risk, ATP III (2002) sex-specific Cox model\par} & {\raggedright replaced by guideline; prognosis\par} & {\raggedright \emph{ACC/\allowbreak{}AHA, US, 2013; AHA, US, 2023.} US guidelines replaced Framingham hard CHD risk with the Pooled Cohort Equations (2013) and then the PREVENT equations (2023) for primary-prevention risk estimation. \cite{goff2013pce,khan2024prevent}\par} \\
{\raggedright MDRD GFR Equation\par} & {\raggedright MDRD (IDMS, 175) with the 1.212 race and 0.742 sex coefficients\par} & {\raggedright replaced by guideline; staging\par} & {\raggedright \emph{NKF-ASN task force, US, 2021.} The 2021 NKF-ASN task force recommends the race-free CKD-EPI 2021 equation and the withdrawal of race-coefficient equations from US laboratory reporting. \cite{delgado2022gfr,inker2021gfr}\par} \\
{\raggedright MELD Na (UNOS/OPTN)\par} & {\raggedright MELD-Na (UNOS 2016): MELD(i) with the sodium term, sodium bounded to 125 to 137, capped at 40\par} & {\raggedright replaced by guideline; allocation\par} & {\raggedright \emph{OPTN/\allowbreak{}UNOS, US, 2023.} OPTN replaced MELD-Na with MELD 3.0 for US liver allocation in July 2023; MELD-Na remains in use outside US allocation. \cite{kim2021meld3}\par} \\
{\raggedright SIRS Criteria\par} & {\raggedright SIRS (1992) four-criteria count\par} & {\raggedright replaced by guideline; diagnosis\par} & {\raggedright \emph{SCCM/\allowbreak{}ESICM Sepsis-3 task force, international, 2016.} Sepsis-3 removed SIRS from the consensus definition of sepsis in favour of SOFA; the 2021 Surviving Sepsis Campaign still lists SIRS among acceptable screening tools. \cite{singer2016sepsis3,evans2021ssc}\par} \\
{\raggedright CHA2DS2-VASc Score for Atrial Fibrillation Stroke Risk\par} & {\raggedright CHA2DS2-VASc (2010) point table with the female-sex point\par} & {\raggedright jurisdiction dependent; prognosis\par} & {\raggedright \emph{ESC/\allowbreak{}EACTS, Europe, 2024; ACC/\allowbreak{}AHA/\allowbreak{}ACCP/\allowbreak{}HRS, US, 2023.} The 2024 ESC guideline moves to CHA2DS2-VA, dropping the sex criterion, while the 2023 ACC/AHA/ACCP/HRS guideline retains CHA2DS2-VASc, so the benchmark's version is current in the US and superseded in Europe. \cite{vangelder2024esc,joglar2024acc}\par} \\
{\raggedright Calcium Correction for Hypoalbuminemia\par} & {\raggedright Simplified Payne: Ca mg/dL + 0.8 x (4.0 - albumin g/dL)\par} & {\raggedright use specific caveat; diagnosis\par} & {\raggedright In 22,658 Alberta patients with paired measurements (2013 to 2019) Payne-adjusted calcium agreed with ionized calcium less well than unadjusted total calcium (63.0 versus 74.5 percent category agreement), with misclassification worst at albumin below 30 g/L, the case the adjustment is meant for. \cite{desgagnes2025calcium,payne1973calcium}\par} \\
{\raggedright Creatinine Clearance (Cockcroft-Gault Equation)\par} & {\raggedright Cockcroft-Gault (1976): (140 - age) x weight x (0.85 if female) / (72 x Scr), weight input chosen by BMI class (actual, ideal or adjusted)\par} & {\raggedright use specific caveat; dosing\par} & {\raggedright \emph{NKF-ASN task force, US, 2021.} No consensus fixes the weight input (actual, ideal or adjusted), which the benchmark selects by BMI class; the 2021 NKF-ASN task force recommends CKD-EPI 2021 for GFR estimation while Cockcroft-Gault persists in drug labelling. \cite{delgado2022gfr}\par} \\
{\raggedright LDL Calculated\par} & {\raggedright Friedewald: total cholesterol - HDL - triglycerides / 5 (mg/dL)\par} & {\raggedright use specific caveat; diagnosis\par} & {\raggedright \emph{AHA/\allowbreak{}ACC, US, 2018.} The 2018 AHA/ACC cholesterol guideline deems direct measurement or a modified estimate (Martin/Hopkins) reasonable when LDL-C is below 70 mg/dL, and Friedewald is invalid at triglycerides of 400 mg/dL or more; the Sampson (NIH) 2020 equation is a newer alternative. \cite{grundy2019cholesterol,martin2013ldl,sampson2020ldl}\par} \\
{\raggedright QTc Bazett Calculator\par} & {\raggedright QT / sqrt(RR)\par} & {\raggedright use specific caveat; diagnosis\par} & {\raggedright \emph{AHA/\allowbreak{}ACCF/\allowbreak{}HRS, US, 2009.} The 2009 AHA/ACCF/HRS ECG statement notes that Bazett over-corrects at high and under-corrects at low heart rates and favours linear formulas; a 2016 head-to-head comparison ranked Fridericia and Framingham best. \cite{rautaharju2009qt,vandenberk2016qtc}\par} \\
{\raggedright Adjusted Body Weight\par} & {\raggedright ABW = IBW + 0.4 x (actual weight - IBW), IBW by Devine\par} & {\raggedright contested coefficient; dosing\par} & {\raggedright The 0.4 correction factor is a dosing convention from aminoglycoside pharmacokinetics with no single derivation, and the Devine IBW it builds on was never empirically derived. \cite{pai2000ideal}\par} \\
{\raggedright Free Water Deficit\par} & {\raggedright TBW fraction (0.6, 0.5, 0.5, 0.45 by age and sex) x weight x (Na / 140 - 1)\par} & {\raggedright contested coefficient; dosing\par} & {\raggedright The total-body-water fractions and the 140 mmol/L target are textbook conventions rather than measured constants, and published versions differ in the fractions used. \cite{adrogue2000hypernatremia}\par} \\
{\raggedright Ideal Body Weight\par} & {\raggedright Devine: 50 kg (45.5 kg if female) + 2.3 kg per inch over 60 in\par} & {\raggedright contested coefficient; dosing\par} & {\raggedright The Devine constants were never empirically derived and misestimate at height extremes; several competing ideal-weight equations exist. \cite{pai2000ideal}\par} \\
{\raggedright Sodium Correction for Hyperglycemia\par} & {\raggedright Hillier 1999: Na + 0.024 x (glucose mg/dL - 100)\par} & {\raggedright contested coefficient; diagnosis\par} & {\raggedright The Hillier 1999 factor of 2.4 mEq/L per 100 mg/dL glucose competes with the Katz 1973 factor of 1.6 that many references still use. \cite{hillier1999sodium,katz1973sodium}\par} \\
{\raggedright APACHE II Score\par} & {\raggedright APACHE II (1985): 12 acute physiology variables plus age and chronic health points\par} & {\raggedright newer version exists; prognosis\par} & {\raggedright APACHE IV (2006) re-estimated the mortality model on a modern ICU cohort; APACHE II remains in wide use and no guideline withdraws it. \cite{zimmerman2006apache}\par} \\
{\raggedright Caprini Score for Venous Thromboembolism (2005)\par} & {\raggedright Caprini 2005 point table\par} & {\raggedright newer version exists; prognosis\par} & {\raggedright A 2013 revision of the Caprini risk assessment model re-weights several items; the 2005 table remains in guideline and calculator use and no body has withdrawn it. \cite{caprini2005thrombosis,cronin2019caprini}\par} \\
{\raggedright Charlson Comorbidity Index (CCI)\par} & {\raggedright Charlson 1987 weights for 17 conditions plus age points\par} & {\raggedright newer version exists; prognosis\par} & {\raggedright Quan 2011 re-weighted the index on six-country data, dropping peptic ulcer disease and changing several weights; the original weights remain in wide use. \cite{quan2011charlson}\par} \\
\addlinespace
\multicolumn{4}{@{}p{\linewidth}@{}}{\emph{No issue identified (39), by intended use:} diagnosis 15, prognosis 8, screening 4, staging 3, dating 3, dosing 2, arithmetic identity 2, conversion 2. Names in the audit file in the repository.} \\
\bottomrule
\end{tabular*}
\caption{Preliminary single-annotator literature audit of the 55 calculators: the 16 with a concern, with formula, type and use, guideline body and date, concern and sources; the 39 others counted by use. Gold answers are the calculator's own output, so a replaced formula still scores correct.}
\label{tab:s9-audit}
\end{table}

\begin{table}[!htbp]
\centering
\footnotesize
\setlength{\tabcolsep}{4pt}
\begin{tabular*}{\linewidth}{@{\extracolsep{\fill}}lrrrrr@{}}
\toprule
\multicolumn{2}{c}{} & \multicolumn{2}{c}{\textbf{Qwen2.5-7B}} & \multicolumn{2}{c}{\textbf{Qwen2.5-32B-AWQ}} \\
\cmidrule(lr){3-4} \cmidrule(lr){5-6}
\textbf{Calculator set} & \textbf{Calcs} & \textbf{pp} & \textbf{Cluster CI} & \textbf{pp} & \textbf{Cluster CI} \\
\midrule
\multicolumn{6}{@{}l}{\emph{Program-Solve vs Open-book arithmetic, formula and gold variables}} \\
\hspace{1em}all & 55 & +3.29 & $[-3.49,\,10.38]$ & +7.05 & $[0.47,\,14.60]$ \\
\hspace{1em}not replaced & 51 & +0.20 & $[-5.82,\,6.41]$ & +3.41 & $[-2.12,\,9.78]$ \\
\hspace{1em}no issue identified & 39 & -0.69 & $[-7.59,\,6.54]$ & +4.36 & $[-2.44,\,12.49]$ \\
\addlinespace
\multicolumn{6}{@{}l}{\emph{Program-Solve vs Gold-Solve library, gold variables}} \\
\hspace{1em}all & 55 & +35.31 & $[21.87,\,48.11]$ & +50.53 & $[38.00,\,62.80]$ \\
\hspace{1em}not replaced & 51 & +32.51 & $[18.63,\,46.10]$ & +47.43 & $[34.88,\,60.25]$ \\
\hspace{1em}no issue identified & 39 & +32.26 & $[15.51,\,48.21]$ & +47.59 & $[32.97,\,61.79]$ \\
\addlinespace
\multicolumn{6}{@{}l}{\emph{Program-Solve + syntax vs Gold-Solve library, gold variables}} \\
\hspace{1em}all & 55 & +37.95 & $[25.24,\,50.38]$ & +50.96 & $[38.47,\,63.09]$ \\
\hspace{1em}not replaced & 51 & +35.51 & $[22.41,\,48.57]$ & +48.27 & $[35.69,\,60.96]$ \\
\hspace{1em}no issue identified & 39 & +36.51 & $[21.20,\,51.23]$ & +48.26 & $[33.82,\,62.28]$ \\
\addlinespace
\multicolumn{6}{@{}l}{\emph{Blind Program-Solve vs Extract-Solve library, both extract}} \\
\hspace{1em}all & 55 & -10.80 & $[-24.35,\,2.64]$ & +5.71 & $[-7.96,\,18.87]$ \\
\hspace{1em}not replaced & 51 & -13.43 & $[-27.65,\,0.77]$ & +4.20 & $[-9.94,\,18.31]$ \\
\hspace{1em}no issue identified & 39 & -14.95 & $[-31.41,\,0.56]$ & +2.87 & $[-13.38,\,18.85]$ \\
\bottomrule
\end{tabular*}
\caption{Exploratory: the four paired gaps on three calculator sets from Table~\ref{tab:s9-audit}: all 55, 51 not replaced by a guideline, 39 with no issue. Percentage points, five seeds, calculator-cluster bootstrap; Gold-Solve abstains outside its 440 cases.}
\label{tab:s10-audit-gaps}
\end{table}

\begin{table}[!htbp]
\centering
\footnotesize
\setlength{\tabcolsep}{4pt}
\begin{tabular*}{\linewidth}{@{\extracolsep{\fill}}lrrrrrrr@{}}
\toprule
\multicolumn{2}{c}{} & \multicolumn{3}{c}{\textbf{Qwen2.5-7B}} & \multicolumn{3}{c}{\textbf{Qwen2.5-32B-AWQ}} \\
\cmidrule(lr){3-5} \cmidrule(lr){6-8}
\textbf{Group} & \textbf{Calcs} & \textbf{Arith.} & \textbf{Prog.} & \textbf{Blind} & \textbf{Arith.} & \textbf{Prog.} & \textbf{Blind} \\
\midrule
Replaced by guideline & 4 & 28.25 & 71.00 & 22.75 & 36.50 & 90.00 & 25.00 \\
Other concern & 12 & 71.92 & 75.00 & 32.58 & 88.25 & 88.58 & 48.50 \\
No issue identified & 39 & 76.54 & 75.85 & 27.18 & 86.82 & 91.18 & 45.56 \\
\bottomrule
\end{tabular*}
\caption{Exploratory: accuracy by audit group, full set, five seeds, for Open-book arithmetic (Arith.), Program-Solve (Prog.) and Blind Program-Solve (Blind). Groups follow Table~\ref{tab:s9-audit}: replaced by a guideline, any other concern, and no issue identified. The replaced group is 4 calculators, so its numbers are indicative only.}
\label{tab:s11-audit-tiers}
\end{table}

\begin{table}[!htbp]
\centering
\scriptsize
\setlength{\tabcolsep}{3pt}
\begin{tabular*}{\linewidth}{@{\extracolsep{\fill}}lrrrrrrrrrrr@{}}
\toprule
\multicolumn{1}{c}{} & \multicolumn{3}{c}{\textbf{Supported 440}} & \multicolumn{3}{c}{\textbf{Unsupported 660}} & \multicolumn{5}{c}{\textbf{Full 1,100}} \\
\cmidrule(lr){2-4} \cmidrule(lr){5-7} \cmidrule(lr){8-12}
\textbf{Arm} & \textbf{Right} & \textbf{Wrong} & \textbf{None} & \textbf{Right} & \textbf{Wrong} & \textbf{None} & \textbf{Right} & \textbf{Wrong} & \textbf{No out.} & \textbf{Exec.} & \textbf{Unread.} \\
\midrule
\multicolumn{12}{@{}l}{\emph{Qwen2.5-7B}} \\
Open-book arithmetic & 83.6 & 16.4 & 0.0 & 64.3 & 35.7 & 0.0 & 72.0 & 28.0 & 0.0 & 0.0 & 0.0 \\
Gold-Solve library & 100.0 & 0.0 & 0.0 & 0.0 & 0.0 & 100.0 & 40.0 & 0.0 & 60.0 & 0.0 & 0.0 \\
Extract-Solve library & 97.1 & 1.4 & 1.6 & 0.0 & 0.0 & 100.0 & 38.8 & 0.6 & 60.6 & 0.0 & 0.0 \\
Program-Solve & 84.2 & 6.8 & 9.0 & 69.4 & 25.5 & 5.1 & 75.3 & 18.0 & 1.3 & 5.3 & 0.1 \\
Program-Solve + syntax & 87.3 & 10.2 & 2.5 & 71.7 & 22.7 & 5.6 & 78.0 & 17.7 & 0.0 & 4.3 & 0.0 \\
Blind Program-Solve & 39.6 & 52.1 & 8.3 & 20.3 & 62.7 & 17.0 & 28.0 & 58.5 & 7.0 & 5.3 & 1.2 \\
\addlinespace
\multicolumn{12}{@{}l}{\emph{Qwen2.5-32B-AWQ}} \\
Open-book arithmetic & 91.5 & 8.6 & 0.0 & 78.2 & 21.9 & 0.0 & 83.5 & 16.5 & 0.0 & 0.0 & 0.0 \\
Gold-Solve library & 100.0 & 0.0 & 0.0 & 0.0 & 0.0 & 100.0 & 40.0 & 0.0 & 60.0 & 0.0 & 0.0 \\
Extract-Solve library & 97.5 & 0.5 & 2.0 & 0.0 & 0.0 & 100.0 & 39.0 & 0.2 & 60.8 & 0.0 & 0.0 \\
Program-Solve & 98.6 & 0.2 & 1.1 & 85.1 & 14.6 & 0.3 & 90.5 & 8.8 & 0.0 & 0.6 & 0.0 \\
Program-Solve + syntax & 98.5 & 0.5 & 1.1 & 86.0 & 13.6 & 0.4 & 91.0 & 8.4 & 0.0 & 0.7 & 0.0 \\
Blind Program-Solve & 57.5 & 41.2 & 1.2 & 36.1 & 60.2 & 3.6 & 44.7 & 52.6 & 0.7 & 1.6 & 0.4 \\
\bottomrule
\end{tabular*}
\caption{Outcome of every case-seed evaluation, in percent, by whether the case's calculator is one of the 22 the library implements. None is no valid answer; the full-set block splits it into no output, an executor failure, and a returned value the scorer could not read. Five seeds pooled.}
\label{tab:s12-breakdown}
\end{table}

\begin{table}[!htbp]
\centering
\scriptsize
\setlength{\tabcolsep}{4pt}
\begin{tabular*}{\linewidth}{@{\extracolsep{\fill}}lrrrrr@{}}
\toprule
\textbf{Reference} & \textbf{Acc.} & \textbf{Gap, pp} & \textbf{Case CI} & \textbf{Cluster CI} & \textbf{Perm.\ $p$} \\
\midrule
\multicolumn{6}{@{}l}{\emph{Qwen2.5-7B, Gold-first, Program-Solve fallback, accuracy 81.64}} \\
Open-book arithmetic & 72.02 & +9.62 & $[6.84,\,12.40]$ & $[2.78,\,17.40]$ & $<$0.001 \\
Program-Solve & 75.31 & +6.33 & $[5.00,\,7.73]$ & $[2.13,\,11.93]$ & $<$0.001 \\
Gold-Solve library & 40.00 & +41.64 & $[38.78,\,44.49]$ & $[30.98,\,52.16]$ & $<$0.001 \\
\addlinespace
\multicolumn{6}{@{}l}{\emph{Qwen2.5-7B, Gold-first, arithmetic fallback, accuracy 78.56}} \\
Open-book arithmetic & 72.02 & +6.55 & $[5.16,\,7.98]$ & $[2.27,\,11.98]$ & $<$0.001 \\
Program-Solve & 75.31 & +3.25 & $[0.38,\,6.09]$ & $[-4.45,\,11.18]$ & 0.024 \\
Gold-Solve library & 40.00 & +38.56 & $[35.76,\,41.31]$ & $[28.36,\,48.96]$ & $<$0.001 \\
\addlinespace
\multicolumn{6}{@{}l}{\emph{Qwen2.5-7B, Extract-first, Program-Solve fallback, accuracy 51.58}} \\
Open-book arithmetic & 72.02 & -20.44 & $[-23.55,\,-17.25]$ & $[-31.15,\,-9.76]$ & $<$0.001 \\
Program-Solve & 75.31 & -23.73 & $[-27.11,\,-20.31]$ & $[-35.35,\,-11.65]$ & $<$0.001 \\
Blind Program-Solve & 28.04 & +23.55 & $[21.05,\,26.07]$ & $[13.98,\,33.95]$ & $<$0.001 \\
Extract-Solve library & 38.84 & +12.75 & $[10.95,\,14.64]$ & $[6.80,\,19.87]$ & $<$0.001 \\
Extract-Solve library, recalled & 46.87 & +4.71 & $[2.93,\,6.56]$ & $[-0.04,\,10.27]$ & $<$0.001 \\
\addlinespace
\multicolumn{6}{@{}l}{\emph{Qwen2.5-7B, Extract-first, arithmetic fallback, accuracy 78.02}} \\
Open-book arithmetic & 72.02 & +6.00 & $[4.62,\,7.44]$ & $[1.76,\,11.35]$ & $<$0.001 \\
Blind Program-Solve & 28.04 & +49.98 & $[47.02,\,52.84]$ & $[40.84,\,59.15]$ & $<$0.001 \\
Extract-Solve library & 38.84 & +39.18 & $[36.45,\,41.98]$ & $[29.09,\,49.42]$ & $<$0.001 \\
\addlinespace
\multicolumn{6}{@{}l}{\emph{Qwen2.5-32B-AWQ, Gold-first, Program-Solve fallback, accuracy 91.07}} \\
Open-book arithmetic & 83.47 & +7.60 & $[5.27,\,9.98]$ & $[0.93,\,15.31]$ & $<$0.001 \\
Program-Solve & 90.53 & +0.55 & $[0.18,\,1.00]$ & $[0.09,\,1.27]$ & 0.032 \\
Gold-Solve library & 40.00 & +51.07 & $[48.05,\,54.02]$ & $[38.78,\,63.09]$ & $<$0.001 \\
\addlinespace
\multicolumn{6}{@{}l}{\emph{Qwen2.5-32B-AWQ, Gold-first, arithmetic fallback, accuracy 86.89}} \\
Open-book arithmetic & 83.47 & +3.42 & $[2.42,\,4.53]$ & $[0.40,\,7.76]$ & $<$0.001 \\
Program-Solve & 90.53 & -3.64 & $[-5.85,\,-1.49]$ & $[-10.49,\,2.20]$ & $<$0.001 \\
Gold-Solve library & 40.00 & +46.89 & $[43.91,\,49.89]$ & $[35.29,\,58.53]$ & $<$0.001 \\
\addlinespace
\multicolumn{6}{@{}l}{\emph{Qwen2.5-32B-AWQ, Extract-first, Program-Solve fallback, accuracy 60.87}} \\
Open-book arithmetic & 83.47 & -22.60 & $[-25.71,\,-19.47]$ & $[-32.15,\,-13.05]$ & $<$0.001 \\
Program-Solve & 90.53 & -29.65 & $[-32.71,\,-26.69]$ & $[-39.42,\,-20.22]$ & $<$0.001 \\
Blind Program-Solve & 44.71 & +16.16 & $[14.02,\,18.36]$ & $[8.35,\,24.87]$ & $<$0.001 \\
Extract-Solve library & 39.00 & +21.87 & $[19.45,\,24.29]$ & $[14.16,\,30.25]$ & $<$0.001 \\
Extract-Solve library, recalled & 50.27 & +10.60 & $[8.20,\,13.04]$ & $[4.89,\,17.25]$ & $<$0.001 \\
\addlinespace
\multicolumn{6}{@{}l}{\emph{Qwen2.5-32B-AWQ, Extract-first, arithmetic fallback, accuracy 86.71}} \\
Open-book arithmetic & 83.47 & +3.24 & $[2.22,\,4.35]$ & $[0.31,\,7.56]$ & $<$0.001 \\
Blind Program-Solve & 44.71 & +42.00 & $[38.85,\,45.09]$ & $[32.85,\,51.49]$ & $<$0.001 \\
Extract-Solve library & 39.00 & +47.71 & $[44.71,\,50.69]$ & $[36.31,\,59.27]$ & $<$0.001 \\
\bottomrule
\end{tabular*}
\caption{Library-first baselines assembled per case and seed. Gold-first answers with the Gold-Solve library on its 440 cases; Extract-first with the Extract-Solve library whenever it answers. Each comes with a Program-Solve fallback and an Open-book arithmetic fallback for the rest. Full 1,100, five seeds; gap is baseline minus reference.}
\label{tab:s13-hybrids}
\end{table}

\begin{table}[!htbp]
\centering
\scriptsize
\setlength{\tabcolsep}{4pt}
\begin{tabular*}{\linewidth}{@{\extracolsep{\fill}}p{5.4cm}p{8.2cm}@{}}
\toprule
{\raggedright \textbf{Setting}\par} & {\raggedright \textbf{Value}\par} \\
\midrule
{\raggedright Model (Hugging Face id, served by vLLM)\par} & {\raggedright Qwen2.5-7B: Qwen/\allowbreak Qwen2.5-7B-Instruct\newline Qwen2.5-32B-AWQ: Qwen/\allowbreak Qwen2.5-32B-Instruct-AWQ\newline Mistral-7B-v0.3: mistralai/\allowbreak Mistral-7B-Instruct-v0.3\newline Phi-3.5-mini: microsoft/\allowbreak Phi-3.5-mini-instruct\par} \\
{\raggedright Checkpoint revision, Hugging Face snapshot (scope: rerun of the formula-reading arms)\par} & {\raggedright Qwen2.5-7B: a09a35458c702b33eeacc393d103063234e8bc28\newline Qwen2.5-32B-AWQ: 5c7cb76a268fc6cfbb9c4777eb24ba6e27f9ee6c\newline Mistral-7B-v0.3: c170c708c41dac9275d15a8fff4eca08d52bab71\newline Phi-3.5-mini: 2fe192450127e6a83f7441aef6e3ca586c338b77\par} \\
{\raggedright Inference server\par} & {\raggedright Qwen2.5-7B: vLLM not recorded, one OpenAI-compatible chat request per turn\newline Qwen2.5-32B-AWQ: vLLM 0.28.0, one OpenAI-compatible chat request per turn\newline Mistral-7B-v0.3: vLLM 0.28.0, one OpenAI-compatible chat request per turn\newline Phi-3.5-mini: vLLM 0.28.0, one OpenAI-compatible chat request per turn\par} \\
{\raggedright Precision\par} & {\raggedright Qwen2.5-7B: bfloat16, the checkpoint default under the server flag\newline Qwen2.5-32B-AWQ: float16 with 4-bit AWQ weights, the checkpoint default under the server flag\newline Mistral-7B-v0.3: bfloat16, the checkpoint default under the server flag\newline Phi-3.5-mini: bfloat16, the checkpoint default under the server flag\par} \\
{\raggedright Formula catalogue and prompt-builder version (the code commit each run manifest records)\par} & {\raggedright Qwen2.5-7B: a0e4ab1f85a9\newline Qwen2.5-32B-AWQ: 0bd8a8597c5f\newline Mistral-7B-v0.3: 0bd8a8597c5f, 507ac149db44\newline Phi-3.5-mini: b03103707bb9\par} \\
{\raggedright Scoring\par} & {\raggedright date\_exact, numeric\_interval, numeric\_parse\_failed; a numeric answer is scored inside the benchmark tolerance, a date exactly\par} \\
{\raggedright Prompt templates\par} & {\raggedright the prompt builders in the runner, one per arm, reproduced verbatim by rerunning it\par} \\
{\raggedright torch and transformers versions (client side)\par} & {\raggedright Qwen2.5-7B: 2.11.0 /\allowbreak  4.41.2\newline Qwen2.5-32B-AWQ: 2.11.0, 2.13.0 /\allowbreak  4.41.2, 5.16.1\newline Mistral-7B-v0.3: 2.11.0, 2.13.0 /\allowbreak  4.41.2, 5.16.1\newline Phi-3.5-mini: 2.11.0, 2.13.0 /\allowbreak  4.41.2, 5.16.1\par} \\
{\raggedright Seeds\par} & {\raggedright 42, 43, 44, 45, 46\par} \\
{\raggedright Temperature, single-call arms (Open-book arithmetic, Page without formula, Extract-Solve extraction, Formulate-Solve tree, the program arms)\par} & {\raggedright 0.0\par} \\
{\raggedright Temperature, sampled candidates (recalled Extract-Solve, three votes; Blind + vote, five programs)\par} & {\raggedright 0.7\par} \\
{\raggedright top-p\par} & {\raggedright not set (server default 1.0)\par} \\
{\raggedright Max output tokens: note-reading arms (Note only, Note only with gold variables, Open-book arithmetic, Page without formula, recalled Extract-Solve candidates)\par} & {\raggedright 4096\par} \\
{\raggedright Max output tokens: extraction call (Extract-Solve library, recalled Extract-Solve delegation)\par} & {\raggedright 512\par} \\
{\raggedright Max output tokens: program arms (Program-Solve and + syntax; Blind Program-Solve and its levers)\par} & {\raggedright 2048; 1024\par} \\
{\raggedright Note budget: Page without formula\par} & {\raggedright 250 words\par} \\
{\raggedright Note budget: note-reading arms with a worked example (Note only, Note only with gold variables, recalled Extract-Solve candidates)\par} & {\raggedright 120 words; the worked example 24 words\par} \\
{\raggedright Note budget: Open-book arithmetic, Extract-Solve extraction, Formulate-Solve tree and every program arm\par} & {\raggedright the whole note, no cap (cap 0)\par} \\
{\raggedright One-shot example\par} & {\raggedright one benchmark case as the worked example; the same case for every seed, except in the two Note only arms, where it changes with the seed\par} \\
\bottomrule
\end{tabular*}
\caption{Reproducibility ledger, run settings, from the run manifests. One value spans the four families unless the row lists them; the model row names the Hugging Face id vLLM served. Every arm makes one OpenAI-compatible chat request per turn, setting only temperature and the output-token cap.}
\label{tab:s14-settings}
\end{table}

\begin{table}[!htbp]
\centering
\scriptsize
\setlength{\tabcolsep}{4pt}
\begin{tabular*}{\linewidth}{@{\extracolsep{\fill}}p{5.4cm}p{8.2cm}@{}}
\toprule
{\raggedright \textbf{Setting}\par} & {\raggedright \textbf{Value}\par} \\
\midrule
{\raggedright Executor: memory cap\par} & {\raggedright 256 MB\par} \\
{\raggedright Executor: CPU time cap\par} & {\raggedright 5 s\par} \\
{\raggedright Executor: wall-clock cap\par} & {\raggedright 5 s\par} \\
{\raggedright Executor: executed-line cap\par} & {\raggedright 100,000 lines\par} \\
{\raggedright Executor: source length cap\par} & {\raggedright 8,000 characters\par} \\
{\raggedright Executor: importable modules\par} & {\raggedright the standard calendar, datetime, math, time modules only\par} \\
{\raggedright Executor: builtins allow-list\par} & {\raggedright a short allow-list of arithmetic, container and type helpers; six exception types; print, whose output is discarded\par} \\
{\raggedright Executor: rejected at parse time\par} & {\raggedright dynamic evaluation, file, attribute and global access, string formatting, async and generator syntax, and any name or attribute starting with an underscore\par} \\
{\raggedright Executor: interpreter\par} & {\raggedright a fresh isolated interpreter process per program; return value only, printed output ignored\par} \\
\bottomrule
\end{tabular*}
\caption{Reproducibility ledger, executor limits. The executor runs each program in a fresh isolated interpreter process under the caps and allow-lists listed; only the return value is read.}
\label{tab:s14b-executor}
\end{table}

\begin{table}[!htbp]
\centering
\scriptsize
\setlength{\tabcolsep}{4pt}
\begin{tabular*}{\linewidth}{@{\extracolsep{\fill}}lrrrrrr@{}}
\toprule
\multicolumn{2}{c}{} & \multicolumn{5}{c}{\textbf{Category, case-seed rows}} \\
\cmidrule(lr){3-7}
\textbf{Set} & \textbf{Rows} & \textbf{Needs reading} & \textbf{Invalid code} & \textbf{Execution failure} & \textbf{Answer-format failure} & \textbf{Correct} \\
\midrule
\multicolumn{7}{@{}l}{\emph{Qwen2.5-7B}} \\
Supported 440 & 2,200 & 150 & 0 & 198 & 0 & 1,852 \\
Unsupported 660 & 3,300 & 841 & 117 & 47 & 5 & 2,290 \\
\addlinespace
\multicolumn{7}{@{}l}{\emph{Qwen2.5-32B-AWQ}} \\
Supported 440 & 2,200 & 5 & 0 & 25 & 0 & 2,170 \\
Unsupported 660 & 3,300 & 480 & 6 & 5 & 0 & 2,809 \\
\bottomrule
\end{tabular*}
\caption{Why Program-Solve still fails when the formula and gold variables are given, audited on the corrected runs, five seeds: every case-seed row by category. Supported means the 440 library cases.}
\label{tab:s15-failure-audit}
\end{table}

\begin{table}[!htbp]
\centering
\footnotesize
\setlength{\tabcolsep}{4pt}
\begin{tabular*}{\linewidth}{@{\extracolsep{\fill}}lrr@{}}
\toprule
\textbf{Annotated category} & \textbf{Qwen2.5-7B} & \textbf{Qwen2.5-32B-AWQ} \\
\midrule
Needs reading, sampled rows & 60 & 60 \\
\hspace{1em}Incorrect formula translation & 5 & 4 \\
\hspace{1em}Wrong variables or units & 46 (1) & 32 (8) \\
\hspace{1em}Wrong branch & 9 & 24 \\
\hspace{1em}\emph{of which the supplied formula was underspecified} & 0 & 0 \\
\addlinespace
Invalid code or execution failure, sampled rows & 20 & 20 \\
\hspace{1em}Invalid code & 7 & 5 \\
\hspace{1em}Execution failure & 13 & 15 \\
\bottomrule
\end{tabular*}
\caption{The annotated sample behind the failure audit: a hash-drawn stratified sample of the rows that needed reading and of the error rows, one category each, one annotator; uncertain counts in parentheses and sent to a second reader.}
\label{tab:s15b-failure-sample}
\end{table}

\begin{table}[!htbp]
\centering
\scriptsize
\setlength{\tabcolsep}{4pt}
\begin{tabular*}{\linewidth}{@{\extracolsep{\fill}}llrrrrrr@{}}
\toprule
\multicolumn{2}{c}{} & \multicolumn{3}{c}{\textbf{Accuracy \%, gap pp}} & \multicolumn{3}{c}{} \\
\cmidrule(lr){3-5}
\textbf{Held fixed} & \textbf{Comparison} & \textbf{Left} & \textbf{Right} & \textbf{Gap} & \textbf{Case CI} & \textbf{Cluster CI} & \textbf{Perm.\ $p$} \\
\midrule
\multicolumn{8}{@{}l}{\emph{Qwen2.5-7B}} \\
budget & Gold variables vs none, 250 words & 30.00 & 25.53 & +4.47 & $[2.95,\,6.00]$ & $[1.62,\,7.80]$ & $<$0.001 \\
access & 250 vs 120 words, no variables & 25.53 & 19.62 & +5.91 & $[4.42,\,7.49]$ & $[3.02,\,9.27]$ & $<$0.001 \\
budget & Open-book arithmetic, whole note, vs 250-word one-shot, no variables & 72.02 & 25.53 & +46.49 & $[43.75,\,49.24]$ & $[38.69,\,54.29]$ & $<$0.001 \\
\addlinespace
\multicolumn{8}{@{}l}{\emph{Qwen2.5-32B-AWQ}} \\
budget & Gold variables vs none, 250 words & 48.69 & 40.84 & +7.85 & $[6.02,\,9.69]$ & $[3.87,\,12.62]$ & $<$0.001 \\
access & 250 vs 120 words, no variables & 40.84 & 28.67 & +12.16 & $[10.29,\,14.18]$ & $[8.00,\,16.58]$ & $<$0.001 \\
budget & Open-book arithmetic, whole note, vs 250-word one-shot, no variables & 83.47 & 40.84 & +42.64 & $[39.87,\,45.45]$ & $[34.40,\,51.16]$ & $<$0.001 \\
\addlinespace
\multicolumn{8}{@{}l}{\emph{Mistral-7B-v0.3}} \\
budget & Gold variables vs none, 250 words & 18.33 & 17.49 & +0.84 & $[-0.64,\,2.33]$ & $[-2.31,\,3.75]$ & 0.271 \\
access & 250 vs 120 words, no variables & 17.49 & 13.35 & +4.15 & $[2.89,\,5.45]$ & $[1.87,\,6.78]$ & $<$0.001 \\
budget & Open-book arithmetic, whole note, vs 250-word one-shot, no variables & 39.05 & 17.49 & +21.56 & $[18.65,\,24.47]$ & $[14.18,\,28.96]$ & $<$0.001 \\
\addlinespace
\multicolumn{8}{@{}l}{\emph{Phi-3.5-mini}} \\
budget & Gold variables vs none, 250 words & 27.07 & 22.93 & +4.15 & $[2.64,\,5.67]$ & $[1.55,\,7.05]$ & $<$0.001 \\
access & 250 vs 120 words, no variables & 22.93 & 18.13 & +4.80 & $[3.29,\,6.38]$ & $[2.18,\,7.62]$ & $<$0.001 \\
budget & Open-book arithmetic, whole note, vs 250-word one-shot, no variables & 52.98 & 22.93 & +30.05 & $[27.25,\,32.98]$ & $[22.47,\,37.65]$ & $<$0.001 \\
\bottomrule
\end{tabular*}
\caption{Note budget and variable access: rows one and two of each block move one factor, row three both; gap is left minus right. Note only reads 120 or 250 words, with or without gold variables; the third row of each block instead sets Open-book arithmetic, which always reads the whole note, against that 250-word one-shot arm without gold variables. Full 1,100, five seeds; intervals as in Table~\ref{tab:s5-paired}.}
\label{tab:s16-entitynone}
\end{table}

\begin{table}[!htbp]
\centering
\footnotesize
\setlength{\tabcolsep}{3pt}
\renewcommand{\arraystretch}{0.9}
\begin{tabular*}{\linewidth}{@{\extracolsep{\fill}}p{2.5cm}p{5.4cm}p{5.1cm}@{}}
\toprule
{\raggedright \textbf{Calculator}\par} & {\raggedright \textbf{What was wrong}\par} & {\raggedright \textbf{What changed}\par} \\
\midrule
{\raggedright Albumin Corrected Anion Gap\par} & {\raggedright Aliased onto plain Anion Gap, which sometimes applied the albumin correction to plain Anion Gap cases too and gave this calculator no dedicated text of its own.\par} & {\raggedright Gave it its own formula text and removed the plain Anion Gap correction sentence.\par} \\
{\raggedright Caprini Score for Venous Thromboembolism (2005)\par} & {\raggedright Surgery-type branch missing; plaster cast given two weights; seven items absent or misweighted.\par} & {\raggedright Replaced with the benchmark's six groups and weights.\par} \\
{\raggedright Creatinine Clearance (Cockcroft-Gault Equation)\par} & {\raggedright Weight chosen by an obesity rule; the benchmark chooses it from body mass index.\par} & {\raggedright Replaced with the three index branches and the ideal weight formula.\par} \\
{\raggedright Framingham Risk Score for Hard Coronary Heart Disease\par} & {\raggedright Described as a points table and lookup; the benchmark runs a sex-specific regression.\par} & {\raggedright Replaced with both coefficient sets, the age caps and the risk conversions.\par} \\
{\raggedright Maintenance Fluids Calculations\par} & {\raggedright States the daily rule divided by twenty-four; the benchmark uses the hourly rule.\par} & {\raggedright Replaced with the hourly rule and its two weight boundaries.\par} \\
{\raggedright PERC Rule for Pulmonary Embolism\par} & {\raggedright Criteria stated in the negative as a rule-out; the benchmark counts the positive criteria.\par} & {\raggedright Restated the criteria as positive and made the answer their count.\par} \\
{\raggedright PSI Score: Pneumonia Severity Index for CAP\par} & {\raggedright Two comorbidity weights too low; the examination and laboratory items named only as groups.\par} & {\raggedright Replaced with all twenty items and their weights.\par} \\
{\raggedright QTc Rautaharju Calculator\par} & {\raggedright Described as a sex-aware correction with unnamed constants; the benchmark uses neither.\par} & {\raggedright Replaced with the benchmark's expression in heart rate.\par} \\
{\raggedright Serum Osmolality\par} & {\raggedright Aliased onto Osmolal Gap, framed around a measured-minus-calculated gap that needs a measured value the note never supplies, instead of the plain calculated value the benchmark scores.\par} & {\raggedright Gave it its own formula text for the plain calculated value.\par} \\
{\raggedright Sodium Correction for Hyperglycemia\par} & {\raggedright Correction constant is 0.016 per unit; the benchmark uses 0.024.\par} & {\raggedright Replaced the constant and named the source.\par} \\
\bottomrule
\end{tabular*}
\caption{Completeness audit of the 55 supplied formula texts against the benchmark's own worked solutions, part one: the 10 texts that could not reproduce the benchmark's number, and the repair. Clinical appropriateness is audited separately.}
\label{tab:s17-catalogue}
\end{table}

\begin{table}[!htbp]
\centering
\footnotesize
\setlength{\tabcolsep}{3pt}
\renewcommand{\arraystretch}{0.9}
\begin{tabular*}{\linewidth}{@{\extracolsep{\fill}}p{2.5cm}p{5.4cm}p{5.1cm}@{}}
\toprule
{\raggedright \textbf{Calculator}\par} & {\raggedright \textbf{What was missing}\par} & {\raggedright \textbf{What changed}\par} \\
\midrule
{\raggedright APACHE II Score\par} & {\raggedright No cutoff for any of the twelve physiology variables; age and chronic health bands misstated; coma scale contribution absent.\par} & {\raggedright Added every band and the coma scale rule.\par} \\
{\raggedright Adjusted Body Weight\par} & {\raggedright Ideal body weight formula and its sex branch absent; adds a precondition the benchmark ignores.\par} & {\raggedright Inlined the two ideal weight formulas, dropped the precondition.\par} \\
{\raggedright Body Surface Area Calculator\par} & {\raggedright Offers a second formula that disagrees, without saying which the benchmark uses.\par} & {\raggedright Kept the root formula the benchmark uses and named it.\par} \\
{\raggedright Charlson Comorbidity Index (CCI)\par} & {\raggedright Age band worth up to four points absent; graded weights flattened.\par} & {\raggedright Added the five age bands and restated every weight as a branch.\par} \\
{\raggedright Estimated Due Date\par} & {\raggedright Cycle-length shift absent.\par} & {\raggedright Added the shift by cycle length minus twenty-eight days.\par} \\
{\raggedright Free Water Deficit\par} & {\raggedright Two of five body water fractions absent, with no age boundary.\par} & {\raggedright Added all five fractions and the age boundaries.\par} \\
{\raggedright Glasgow Coma Score (GCS)\par} & {\raggedright Convention for a component not mentioned or not testable absent.\par} & {\raggedright Added the full-score convention for such components.\par} \\
{\raggedright Glasgow-Blatchford Bleeding Score (GBS)\par} & {\raggedright Point ranges given for urea, hemoglobin and pressure without their thresholds.\par} & {\raggedright Added every threshold for the three graded items.\par} \\
{\raggedright HAS-BLED Score for Major Bleeding Risk\par} & {\raggedright Thresholds for renal, liver, unstable clotting and alcohol absent.\par} & {\raggedright Added the four thresholds.\par} \\
{\raggedright HEART Score for Major Cardiac Events\par} & {\raggedright Components given as ranges without the branch that assigns the value.\par} & {\raggedright Added the branch definitions for all five components.\par} \\
{\raggedright MELD Na (UNOS/OPTN)\par} & {\raggedright Only the sodium adjustment: base equation, bounds, dialysis rule, rounding, gate and cap absent.\par} & {\raggedright Added the bounds, base equation, gate and cap.\par} \\
{\raggedright Morphine Milligram Equivalents (MME) Calculator\par} & {\raggedright None of the eleven conversion factors given.\par} & {\raggedright Added the eleven factors.\par} \\
{\raggedright QTc Bazett Calculator\par} & {\raggedright Does not derive the interval between beats from the heart rate.\par} & {\raggedright Added that convention.\par} \\
{\raggedright QTc Framingham Calculator\par} & {\raggedright Does not derive the interval between beats from the heart rate.\par} & {\raggedright Added that convention.\par} \\
{\raggedright QTc Fridericia Calculator\par} & {\raggedright Does not derive the interval between beats from the heart rate.\par} & {\raggedright Added that convention.\par} \\
{\raggedright Sequential Organ Failure Assessment (SOFA) Score\par} & {\raggedright Six organ systems named with ranges but no cutoffs.\par} & {\raggedright Added every cutoff, with the ventilation and urine output branches.\par} \\
{\raggedright Steroid Conversion Calculator\par} & {\raggedright Three of the eight equivalent doses absent, and no conversion step.\par} & {\raggedright Added the three doses and the conversion step.\par} \\
{\raggedright Target weight\par} & {\raggedright No formula at all, only a pointer to the question.\par} & {\raggedright Replaced with the target index times height squared.\par} \\
\addlinespace
\multicolumn{3}{@{}p{\linewidth}@{}}{\emph{Already reproduced the benchmark (27):} unchanged, so the committed prompts are untouched. Names in the audit file.} \\
\bottomrule
\end{tabular*}
\caption{Completeness audit, part two: the 18 texts that omitted a constant, unit, branch or convention the benchmark applies, and the repair; the 27 already complete are unchanged.}
\label{tab:s17b-catalogue-incomplete}
\end{table}

\begin{table}[!htbp]
\centering
\footnotesize
\setlength{\tabcolsep}{4pt}
\begin{tabular*}{\linewidth}{@{\extracolsep{\fill}}lrrrrrrr@{}}
\toprule
\textbf{Baseline} & \textbf{Library \%} & \textbf{Program} & \textbf{Arith.} & \textbf{Gap, pp} & \textbf{Case CI} & \textbf{Cluster CI} & \textbf{Perm.\ $p$} \\
\midrule
\multicolumn{8}{@{}l}{\emph{Qwen2.5-7B}} \\
Gold-first & 40.0 & 81.64 & 78.56 & +3.07 & $[0.64,\,5.55]$ & $[-2.20,\,9.15]$ & 0.014 \\
Extract-first & 39.4 & 51.58 & 78.02 & -26.44 & $[-29.05,\,-23.78]$ & $[-34.89,\,-18.22]$ & $<$0.001 \\
\addlinespace
\multicolumn{8}{@{}l}{\emph{Qwen2.5-32B-AWQ}} \\
Gold-first & 40.0 & 91.07 & 86.89 & +4.18 & $[2.11,\,6.36]$ & $[-1.60,\,10.96]$ & $<$0.001 \\
Extract-first & 39.2 & 60.87 & 86.71 & -25.84 & $[-28.65,\,-23.04]$ & $[-34.25,\,-17.91]$ & $<$0.001 \\
\bottomrule
\end{tabular*}
\caption{The same library-first baseline with each fallback, on the same cases and seeds: Program-Solve minus Open-book arithmetic. Library share is the percentage of case-seed rows the library answers; the fallback answers the rest. Full 1,100, five seeds, intervals as in Table~\ref{tab:s5-paired}.}
\label{tab:s18-fallbacks}
\end{table}

\begin{table}[!htbp]
\centering
\footnotesize
\setlength{\tabcolsep}{4pt}
\begin{tabular*}{\linewidth}{@{\extracolsep{\fill}}lrrrrrr@{}}
\toprule
\textbf{Arm} & \textbf{Acc.} & \textbf{Seed sd} & \textbf{Agree, \%} & \textbf{Pairwise, \%} & \textbf{Right, \%} & \textbf{Wrong, \%} \\
\midrule
\multicolumn{7}{@{}l}{\emph{Qwen2.5-7B}} \\
Note only$^{\dagger}$ & 19.62 & 1.28 & 71.5 & 86.3 & 8.4 & 63.2 \\
Note only with gold variables$^{\dagger}$ & 31.38 & 1.76 & 59.6 & 80.7 & 14.4 & 45.2 \\
Open-book arithmetic & 72.02 & 0.35 & 90.5 & 95.4 & 67.3 & 23.3 \\
Extract-Solve library & 38.84 & 0.08 & 99.8 & 99.9 & 38.7 & 61.1 \\
Blind Program-Solve & 28.04 & 0.47 & 88.1 & 94.2 & 22.7 & 65.4 \\
Program-Solve & 75.31 & 0.47 & 92.2 & 96.0 & 71.4 & 20.8 \\
Program-Solve + syntax & 77.95 & 0.44 & 92.0 & 95.8 & 73.9 & 18.1 \\
\addlinespace
\multicolumn{7}{@{}l}{\emph{Qwen2.5-32B-AWQ}} \\
Note only$^{\dagger}$ & 28.67 & 1.65 & 75.2 & 88.0 & 16.8 & 58.4 \\
Note only with gold variables$^{\dagger}$ & 48.44 & 1.81 & 58.4 & 79.5 & 26.6 & 31.7 \\
Open-book arithmetic & 83.47 & 0.17 & 98.8 & 99.4 & 83.0 & 15.8 \\
Extract-Solve library & 39.00 & 0.00 & 100.0 & 100.0 & 39.0 & 61.0 \\
Blind Program-Solve & 44.71 & 0.24 & 98.1 & 99.1 & 43.8 & 54.3 \\
Program-Solve & 90.53 & 0.08 & 99.6 & 99.8 & 90.4 & 9.3 \\
Program-Solve + syntax & 90.96 & 0.14 & 99.2 & 99.6 & 90.5 & 8.6 \\
\addlinespace
\multicolumn{7}{@{}l}{\emph{Mistral-7B-v0.3}} \\
Note only$^{\dagger}$ & 13.35 & 0.94 & 66.1 & 83.8 & 1.6 & 64.5 \\
Note only with gold variables$^{\dagger}$ & 18.93 & 1.07 & 55.6 & 78.3 & 2.1 & 53.5 \\
Open-book arithmetic & 39.05 & 0.25 & 91.5 & 96.1 & 35.0 & 56.5 \\
Extract-Solve library & 34.89 & 0.08 & 99.5 & 99.8 & 34.5 & 64.9 \\
Blind Program-Solve & 11.16 & 0.35 & 90.5 & 95.5 & 7.1 & 83.5 \\
Program-Solve & 31.44 & 0.84 & 83.9 & 92.0 & 22.7 & 61.2 \\
Program-Solve + syntax & 26.95 & 0.73 & 89.8 & 95.3 & 21.9 & 67.9 \\
\addlinespace
\multicolumn{7}{@{}l}{\emph{Phi-3.5-mini}} \\
Note only$^{\dagger}$ & 18.13 & 1.01 & 68.3 & 84.3 & 4.9 & 63.4 \\
Note only with gold variables$^{\dagger}$ & 27.07 & 1.53 & 58.5 & 79.7 & 9.0 & 49.5 \\
Open-book arithmetic & 52.98 & 0.59 & 92.3 & 96.5 & 49.1 & 43.2 \\
Extract-Solve library & 35.04 & 0.37 & 95.0 & 97.5 & 32.2 & 62.8 \\
Blind Program-Solve & 23.36 & 0.61 & 78.5 & 89.5 & 13.8 & 64.6 \\
Program-Solve & 57.13 & 0.80 & 88.5 & 95.0 & 50.6 & 37.9 \\
Program-Solve + syntax & 57.25 & 0.28 & 91.8 & 96.5 & 52.8 & 39.0 \\
\bottomrule
\end{tabular*}
\caption{The five seeds fix the case order, the checkpoint name and, in the marked arms, the worked example; no seed reaches the server and every arm here decodes greedily. Agree: cases all five seeds score alike. Pairwise: mean over the ten seed pairs. Right, wrong: unanimous cases.}
\label{tab:s19-seeds}
\end{table}

\end{document}